\documentclass{bmvc2k}
\usepackage[titletoc,title]{appendix}
\usepackage{algorithm}
\usepackage{algorithmicx}
\usepackage[noend]{algpseudocode}
\usepackage{amsfonts}
\usepackage{amsmath}
\usepackage{amssymb}
\usepackage{amstext}
\usepackage{array}
\usepackage{bm}
\usepackage{color}
\usepackage{colortbl}
\usepackage{cleveref}
\usepackage{enumerate}
\usepackage{epsfig}
\usepackage{float}
\usepackage{graphicx}
\usepackage{epstopdf}
\usepackage{mathrsfs}
\usepackage{mathtools}
\usepackage{multirow}
\usepackage{paralist}
\usepackage{pdfpages}
\usepackage{times}
\usepackage{url}
\usepackage{xspace}
\usepackage{hyperref}

\newcommand{\suchthat}{\;\ifnum\currentgrouptype=16 \middle\fi|\;}
\DeclareMathOperator*{\argmin}{argmin}
\DeclareMathOperator*{\argmax}{argmax}

\newcommand{\rootsift}{RootSIFT\xspace}
\newcommand{\rootsifts}{RootSIFTs\xspace}

\newcommand{\inv}{{-1}}

\newcommand{\ve}[1][x]{\ensuremath{\mathbf{#1}}\xspace}

\newcount\colveccount
\newcommand*\colvec[1]{
        \global\colveccount#1
        \begin{pmatrix}
        \colvecnext
}
\def\colvecnext#1{
        #1
        \global\advance\colveccount-1
        \ifnum\colveccount>0
                \\
                \expandafter\colvecnext
        \else
                \end{pmatrix}
        \fi
}

\newtoks\rowvectoks
\newcommand{\rowvec}[2]{%
  \rowvectoks={#2}\count255=#1\relax
  \advance\count255 by -1
  \rowvecnexta}
\newcommand{\rowvecnexta}{%
  \ifnum\count255>0
    \expandafter\rowvecnextb
  \else
    \begin{pmatrix}\the\rowvectoks\end{pmatrix}
  \fi}
\newcommand\rowvecnextb[1]{%
    \rowvectoks=\expandafter{\the\rowvectoks&#1}%
    \advance\count255 by -1
    \rowvecnexta}

\newcommand{\ie}{{\em i.e.}\xspace}
\newcommand{\eg}{{\em e.g.}\xspace}
\newcommand{\etal}{et al.}

\newcommand{\T}{{\!\top}}

\def\oversim#1#2{\lower .5pt\vbox{\lineskiplimit=\maxdimen \lineskip=.5pt
    \ialign{$\mathsurround=0pt #1\hfil ##\hfil $\crcr #2\crcr \sim\crcr }}}

\newcommand{\tribox}[4][.3\linewidth]{\mbox{} \hfill
    \parbox[t]{#1}{\centering #2} \hfill \parbox[t]{#1}{\centering #3} \hfill
    \parbox[t]{#1}{\centering #4} \hfill \mbox{}}

\newcommand{\eq}[1]{Eq.~{#1}}
\newcommand{\fig}[1]{Fig.~{#1}}
\newcommand{\figs}[1]{Figs.~{#1}}

\newcommand{\secref}[1]{section~{#1}}

\definecolor{faded}{rgb}{.97,.97,.97}

\makeatletter
\newcommand*{\pmzeroslash}{%
  \nfss@text{%
    \sbox0{0}%
    \sbox2{/}%
    \sbox4{%
      \raise\dimexpr((\ht0-\dp0)-(\ht2-\dp2))/2\relax\copy2 %
    }%
    \ooalign{%
      \hfill\copy4 \hfill\cr
      \hfill0\hfill\cr
    }%
    \vphantom{0\copy4 }
  }%
}
\makeatother

\renewcommand{\paragraph}[1]{\vspace{.25\baselineskip}\noindent{\bf #1}\xspace}

\newcommand{\inlinelist}[1]{\begin{inparaenum}[(a)] #1 \end{inparaenum}}

\newcommand{\meas}{\ensuremath{\ve[x]}\xspace}
\newcommand{\kps}{\ensuremath{\ve[x]^K}\xspace}
\newcommand{\rgnls}{\ensuremath{\ve[x]^R}\xspace}
\newcommand{\kp}[1]{\ensuremath{\ve[x]^K_{#1}}\xspace}
\newcommand{\rgnl}[1]{\ensuremath{\ve[x]^R_{#1}}\xspace}

\newcommand{\kplblg}[1]{\ensuremath{\ve[y]^K_{#1}}\xspace}
\newcommand{\hkplblg}[1]{\ensuremath{\ve[\hat{y}]^K_{#1}}\xspace}
\newcommand{\rgnlblg}[1]{\ensuremath{\ve[y]^R_{#1}}\xspace}
\newcommand{\hrgnlblg}[1]{\ensuremath{\ve[\hat{y}]^R_{#1}}\xspace}
\newcommand{\kplblgs}[0]{\ensuremath{\ve[y]^K}\xspace}

\newcommand{\lblg}[0]{\ensuremath{\ve[y]}\xspace}

\algrenewcommand\algorithmicrequire{\textbf{Precondition:}}
\algrenewcommand\algorithmicensure{\textbf{Postcondition:}}

\DeclareRobustCommand{\inlinelist}[1]{\begin{inparaenum}[(a)] #1 \end{inparaenum}}

\newcommand\blfootnote[1]{%
  \begingroup
  \renewcommand\thefootnote{}\footnote{#1}%
  \addtocounter{footnote}{-1}%
  \endgroup
}

\newcommand{\ic}[2]{\parbox{#1\linewidth}{\centering \includegraphics[width=0.99\linewidth]{#2}}}

\def\mytitle{Coplanar Repeats by Energy Minimization}

\title{\mytitle}

\addauthor{James Pritts}{http://cmp.felk.cvut.cz/~prittjam}{1}
\addauthor{Denys Rozumnyi}{rozumden@cmp.felk.cvut.cz}{1}
\addauthor{M. Pawan Kumar}{http://mpawankumar.info}{2}
\addauthor{Ondrej Chum}{http://cmp.felk.cvut.cz/~chum}{1}

\addinstitution{
 The Center for Machine Perception \\
 Faculty of Electrical Engineering \\
 Czech Technical University \\
 Prague, CZ
}
\addinstitution{
Department of Engineering Science \\
University\ of Oxford \\
Oxford, UK
}

\runninghead{Pritts, Rozumnyi, Kumar, Chum}{Coplanar Repeats by Eng. Minimization}

\def\eg{\emph{e.g}\bmvaOneDot}

\def\etal{\emph{et al}\bmvaOneDot}

\begin{document}

\maketitle

\begin{abstract} 
This paper proposes an automated method to detect, group and rectify
arbitrarily-arranged coplanar repeated elements via energy
minimization.  The proposed energy functional combines several
features that model how planes with coplanar repeats are projected
into images and captures global interactions between different
coplanar repeat groups and scene planes.  An inference framework based
on a recent variant of $\alpha$-expansion is described and fast
convergence is demonstrated. We compare the proposed method to two
widely-used geometric multi-model fitting methods using a new dataset
of annotated images containing multiple scene planes with coplanar
repeats in varied arrangements. The evaluation shows a significant
improvement in the accuracy of rectifications computed from coplanar
repeats detected with the proposed method versus those detected with
the baseline methods.

\end{abstract}

\section{Introduction} 
The importance of detecting and modeling imaged repeated scene
elements grows with the increasing usage of scene-understanding
systems in urban settings, where man-made objects predominate and
coplanar repeated structures are common. Most state-of-the-art repeat
detection and modeling methods take a greedy approach that follows
appearance-based clustering of extracted keypoints with geometric
verification.  Greedy methods have a common drawback: Sooner or later
the wrong choice will be made in a sequence of threshold tests
resulting in an irrevocable error, which makes a pipeline approach too
fragile for use on large image databases.

We propose a global energy model for grouping coplanar repeats and
scene plane detection. The energy functional combines features
encouraging \begin{inparaenum}[(i)]\item the geometric and appearance
  consistency of coplanar repeated elements, \item the spatial and
  color cohesion of detected scene planes, \item and a parsimonious
  model description of coplanar repeat groups and scene
  planes. \end{inparaenum} The energy is minimzed by block-coordinate
descent, which alternates between grouping extracted keypoints into
coplanar repeats by labeling (see \figs 1,2) and regresses the
continuous parameters that model the geometries and appearances of
coplanar repeat groups and their underlying scene planes. Inference is
fast even for larger problems (see \secref{\ref{sec:Evaluation}}).

Comparison to state-of-the-art coplanar repeat detection methods is
complicated by the fact that many prior methods were either evaluated
on small datasets, include only qualitative results, or were
restricted to images with repeats having a particular symmetry. We
evaluate the proposed method on a new annotated dataset of 113 images.
The images have from 1 to 5 scene planes containing translation,
reflection, or rotation symmetries that repeat periodically or
arbitrarily. Performance is measured by comparing the quality of
rectifications computed from detected coplanar repeat groups versus
rectifications computed from the annotated coplanar repeat groups of
the dataset.

\begin{figure}[t!] \centering
  \begin{tabular}{c@{\hspace{7pt}}c@{\hspace{7pt}}c@{}c|c}
  \ic{0.25}{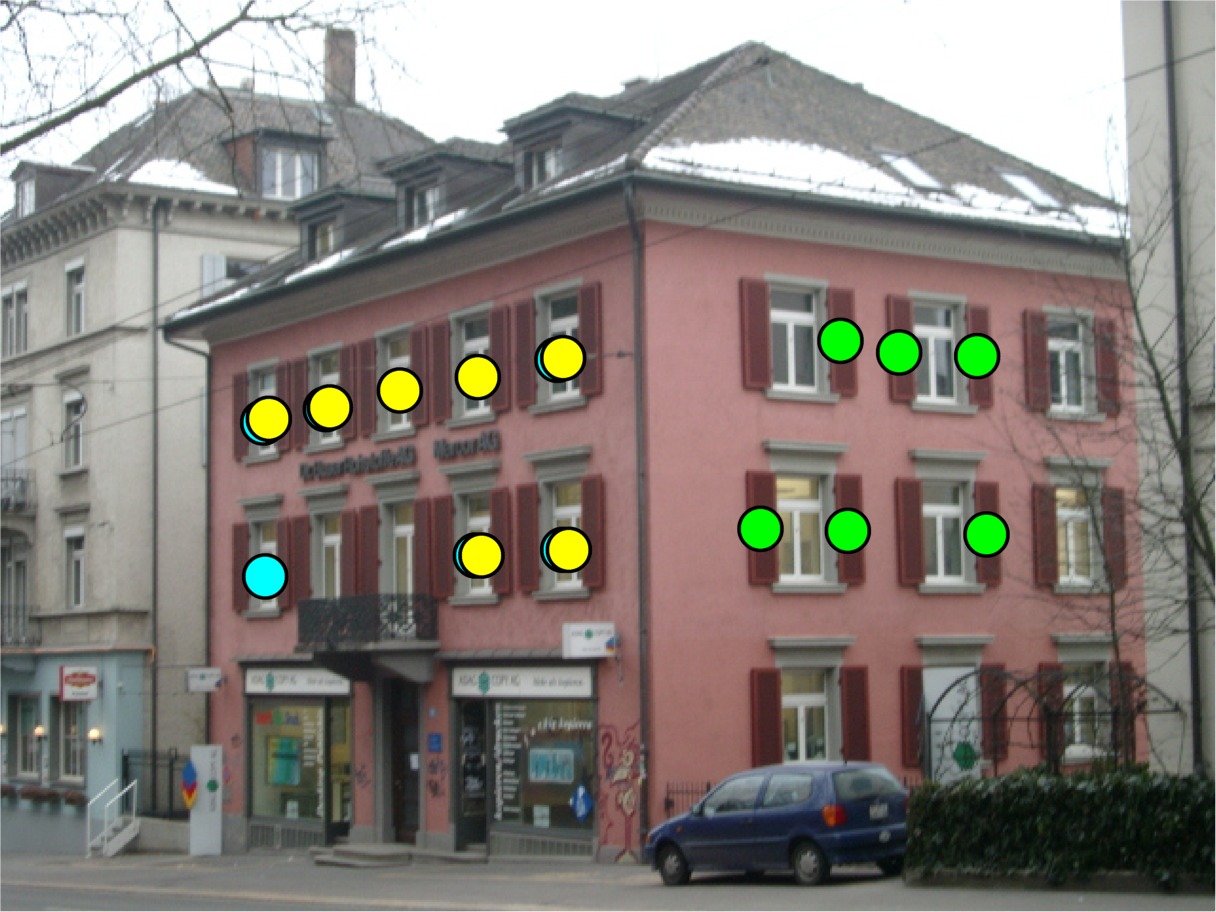} &
  \parbox{.13\linewidth}
  { \centering
  \colorbox{green}
  {\includegraphics[width=7mm]{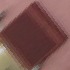} 
  \includegraphics[width=7mm]{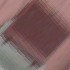}} 

  \colorbox{cyan}{\includegraphics[width=7mm]{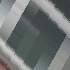} 
  \includegraphics[width=7mm]{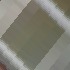}} 

  \colorbox{yellow}
  {\includegraphics[width=7mm]{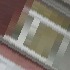}  
  \includegraphics[width=7mm]{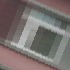}} 
  } 
  &
  \ic{0.25}{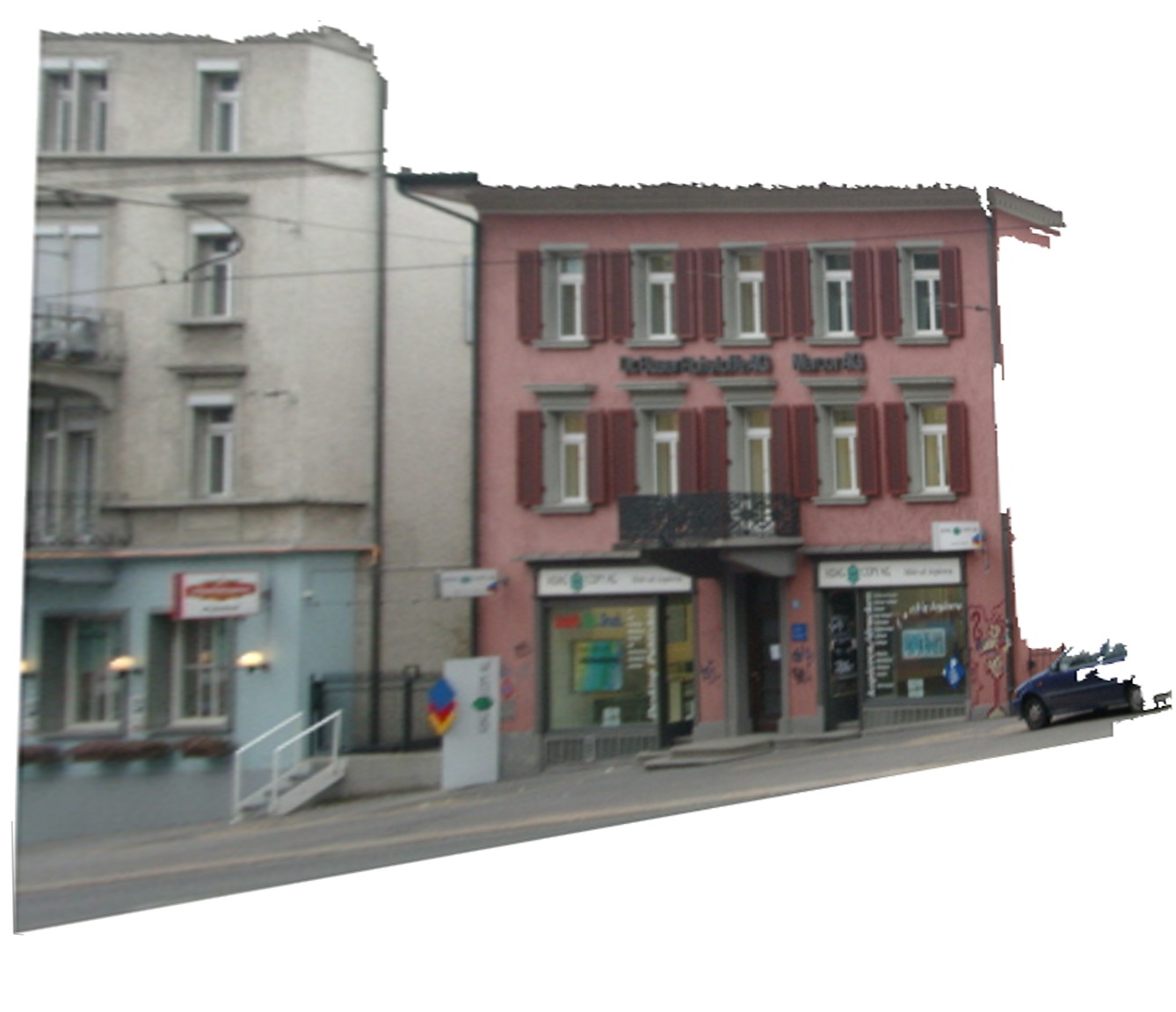} &
  \ic{0.12}{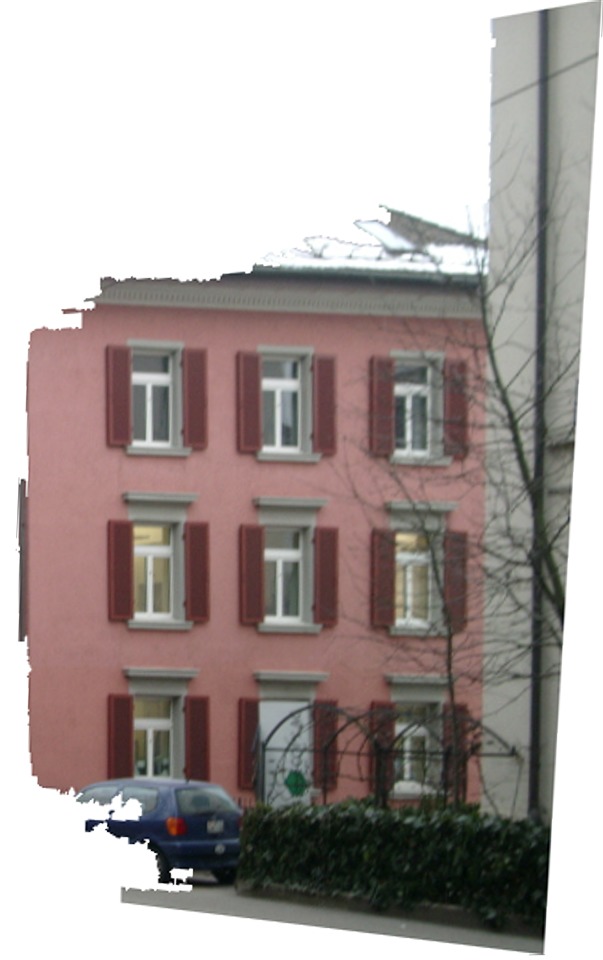} &
   \ic{0.12}{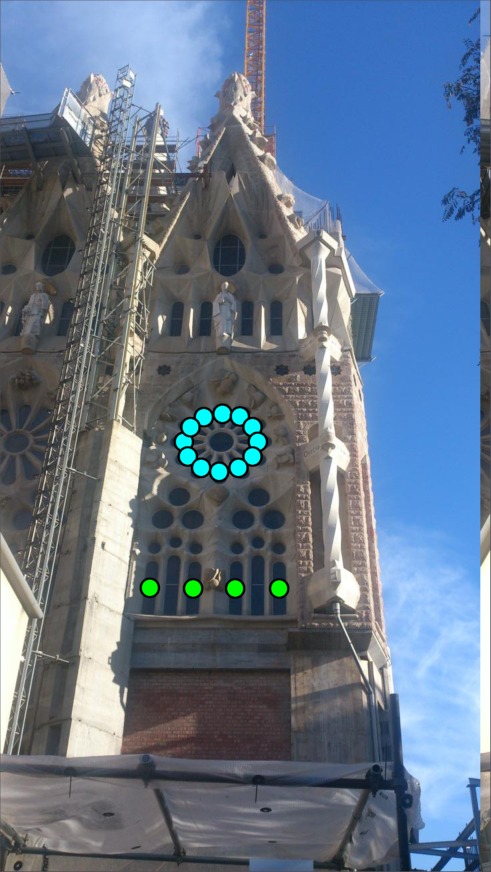} \\
   (a) & (b) & (c) & (d) & (e) \\
\end{tabular}
\caption{Grouping and rectification of coplanar repeats: (a) a subset
  of the detected coplanar repeats is denoted by colored dots, (b)
  rectification of the most distant keypoint pairs grouped as coplanar
  repeats\textemdash repeat group membership is encoded by the colored
  border, (c,d) rectified and segmented scene planes, (e) Translation
  and rotation symmetric keypoints labeled as distinct coplanar
  repeats.}
\end{figure}

\section{Related Work} 
Repeat grouping is a well-studied computer vision task. Many variants
of hypothesize-and-verify pipelines were proposed in the prior
literature for grouping repeats. Typical distinctions between methods
are the variants of used feature types, geometric constraints, and
scene geometry estimators. Two closely related early methods by
Schaffalitzky~\etal~\cite{Schaffalitzky-BMVC98} and
Tuytelaars~\etal~\cite{Tuytelaars-PAMI03} estimate homologies that are
compatible with detected fixed points and lines induced by
periodicities, symmetries and reflections of scene
elements. Liebowitz~\etal~\cite{Liebowitz-CVPR98} use metric
invariants to identify repeats in an affine-rectified frame estimated
from imaged parallel scene lines.

More recent approaches eliminate the need for any scene structure
other than the coplanar repeated scene elements and work for
arbitrarily arranged repeats (\ie, rigidly transformed on the scene
plane). Chum \etal~\cite{Chum-ACCV10} introduce an algebraic
constraint on the local scale change of a planar feature and use it to
verify that tentative repeats have equal scale in a rectified frame
(this constraint is included in the proposed energy function).
Pritts~\etal~\cite{Pritts-CVPR14} introduce constraints specific to
rotated and reflected repeated elements in an affine rectified frame
and generatively build a model of the pattern rectified to within a
similarity of the scene.

Two frequently cited approaches use energy minimization frameworks.
Park \etal~\cite{Park-PAMI09} minimize an energy that measures the
compatibility of a deformable lattice to imaged uniform grids of
repetitions. Wu \etal~\cite{Wu-ECCV10} refine vanishing point
estimates of an imaged building facade by minimizing the difference
between detected symmetries across repetition boundaries of the
facade.

None of the reviewed approaches globally model repeats; rather, there
is an assumption that a dominant plane is present, or repeat grouping
proceeds greedily by detecting scene planes sequentially. A
significant subset of the reviewed literature requires the presence of
special scene structure like parallel scene lines or lattices, which
limits their applicability.

\section{Scene Model}
\label{sec:label_space}
\label{sec:scene_model}
The scene model has three types of outputs: The first output is a
grouping of detected keypoints (see \figs 2a-2d) into coplanar repeats
(see \figs 1a,1e).  Random variables $Y^K$ jointly assign keypoints to
keypoint groups with mutually compatible geometry and appearance and
to planar scene surfaces. Each random variable of $Y^K$ is from the
set $\mathcal{Y}_K = \{\,1 \ldots
N_G,\, \varnothing\,\} \times \{\,1 \dots N_V,b\,\}$. Here $N_G$ is
the number of clusters of keypoints that were grouped based on their
similarity in appearance, and $N_V$ is the estimated number of planar
surfaces in the scene. A particular labeling of $Y^K$ is
denoted \kplblgs. The assignment of the $i$-th keypoint to a
compatible keypoint cluster is indexed as \kplblg{ig}, and its
assignment to a scene plane is indexed as \kplblg{iv}. The empty set
$\varnothing$ is assigned if keypoint $i$ does not repeat,
$\kplblg{ig}=\varnothing$, and the token $b$ is assigned to a keypoint
if it does not lie on a planar surface. Background keypoints cannot be
assigned to a repeat group, so they are assigned the ordered pair
$(\,\varnothing,b \,).$ The non-planar surfaces are collectively
called the background.  The sets of keypoints assigned to the same
keypoint cluster and scene plane are the coplanar repeated patterns.

\begin{table}[t!]
\begin{center}
\begin{tabular}{ |c|c|c|c| } 
 \hline
 Term & Description & Term & Description \\ 
 \hline
 \kp{i} & keypoint, see \fig 2d & $\varnothing$ & keypoint is a singelton  \\  
  \kp{iw} & point of a keypoint & $N_G$ & number of keypoint clusters \\
 $\rgnls_j$ & image region, see \fig 2e & $N_V$ & number of scene planes \\ 
 $\meas$ & all measurements &  $\beta^K(\kplblgs)$ & geom./app. params. for repeats \\ 
 \kplblg{ig} & keypoint $\leftrightarrow$ cluster & $\beta^R(\kplblgs,\ve[y]^R)$ & geom./app. params. for planes \\
 \kplblg{iv} & keypoint $\leftrightarrow$ scene plane & $\beta(\ve[y])$ & joint parameter vector \\
 \kplblg{i} & keypt. label, $(\kplblg{ig},\kplblg{iv})$, see \fig 1a & $\psi(\cdot)$ & joint feature vector \\
 $\ve[y]^R_j$ & region $\leftrightarrow$ scene plane, see \fig 1d & \ve[w] & feature weight vector \\
 $\ve[y]$ & joint labeling & $\ve[l]_n$ & scene plane vanishing line \\
 b & keypt./region is on background & $H_{\ve[l]_n}(\cdot)$ & rectifying transform from $\ve[l]_n$ \\
\hline
\end{tabular}
\caption{The most commonly used scene model denotations.}
\end{center}
\end{table}

The second output is a labeling of image regions as planar surfaces
and background. The image regions are small and connected areas of
similar color that are detected as SEEDS
superpixels \cite{VandenBergh-ECCV12} (see \fig 2e). Random variables
$Y^R$ assign image regions to planar surfaces and the background,
where each random variable of $Y^R$ is from the set $ \mathcal{Y}_R
= \{\,1 \dots N_V,b\,\}$. As before, $N_v$ and $b$ are the estimated
number of planar surfaces and the background token, respectively.  A
particular labeling of $Y^R$ is denoted $\ve[y]^R$, and the labeling
partitions the image regions into larger components that correspond to
contiguous planar surfaces of the scene or background. The assignment
of the $j$-th region to a scene plane or to background is indexed
as \rgnlblg{j}.

The third output is a set of continuous random variables modeling the
geometries and appearances of the sets of coplanar repeats and the
scene planes. The geometries and appearances of coplanar repeats are
functions of the keypoint assignments and are given by the dependent
random variables $B^K(Y^K)$. The corresponding parameter estimates are
denoted as $\beta^K(\kplblgs)$. The geometries and appearances of the
scene planes are functions of $Y^K$ and $Y^R$, and are given by
dependent random variables $B^R(Y^K,Y^R)$. The parameters
$\beta^R(\kplblgs,\ve[y]^R)$ represent the colors of the scene
surfaces and the orientations of scene planes. 

The joint labeling and parameter vector for the entire model are
respectively denoted $\ve[y]={\kplblgs}^\frown{\ve[y]^R}$ and
$\beta(\ve[y])={\beta^K}(\kplblgs)^\frown\beta^R(\kplblgs,\ve[y]^R)$.

\subsection{Energy Function}
The joint feature vector $\psi(\cdot)$ encodes potentials that measure
\begin{inparaenum}[(i)] 
\item coplanar repeats consist of keypoints that have similar appearnace and the same area in the preimage, \item the scene planes
and background should consist of image regions with the same color
distributions, \item surfaces should be contiguous and that nearby
repeated content should be on the same surface, \item and scenes
should have a parsimonious description.
\end{inparaenum}
A minimal energy labeling \lblg and parameter set $\beta(\ve[y])$ are
sought by solving the energy minimization task
\begin{equation}
\label{eq:energy_minimization}
\argmin_{\lblg,\beta} \, \underbrace{\ve[w]^{\T}\psi(\ve[x],\ve[y],\beta(\ve[y]))}_{E \text{ (energy)}},
\end{equation}
where $\ve[x]$ are the detected salient image patches and
over-segmented regions of the image, and $\ve[w]$ is a weight
vector. The components of $\ve[w]$ take on different meanings
depending on their paired features and are discussed
in \Crefrange{sec:unary_features_repeats}{sec:label_subset_costs}.

%

\begin{figure}[t] \centering
\label{fig:keypoint_construction}
\begin{tabular}{c@{\hspace{13pt}}c@{}c@{\hspace{9pt}}c@{\hspace{13pt}}|c@{\hspace{5pt}}c}
\parbox{.138\linewidth}{\centering {\parbox{0.99\linewidth}{\setlength\fboxrule{0.0pt} \fbox{\includegraphics[width=\linewidth,trim={30 30 30 30}]{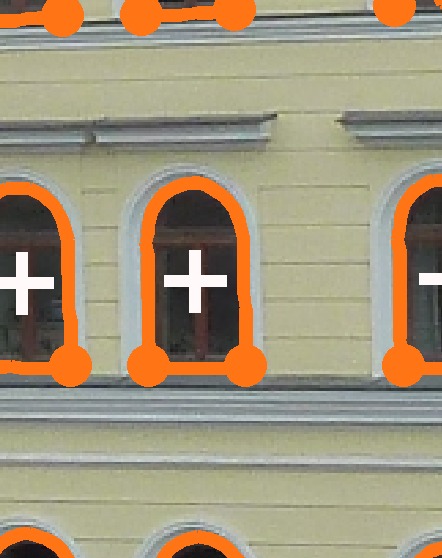}}}}
} &
\parbox{.1\linewidth}
{
\centering
\ic{0.99}{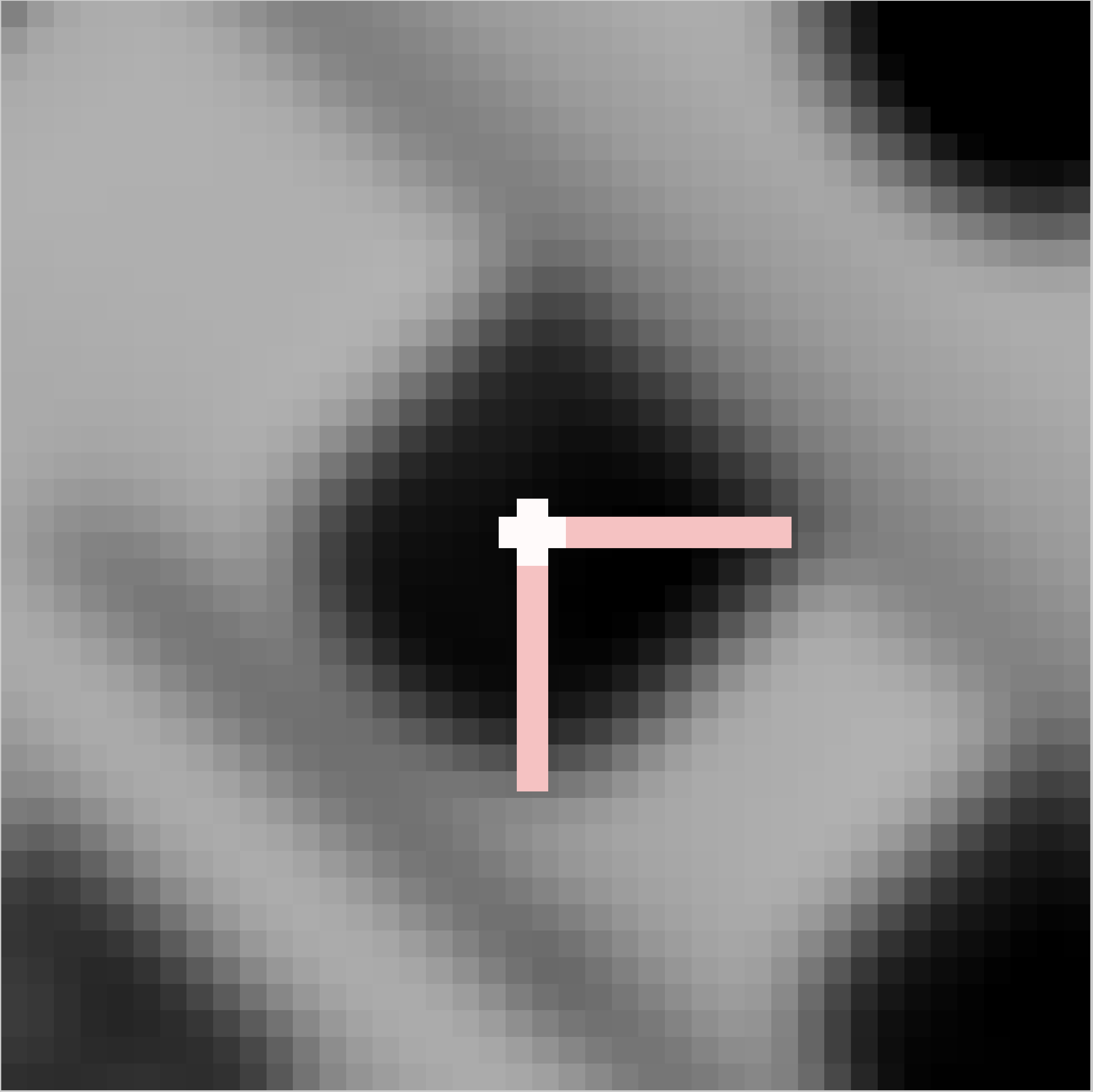}\\ 
\ic{0.99}{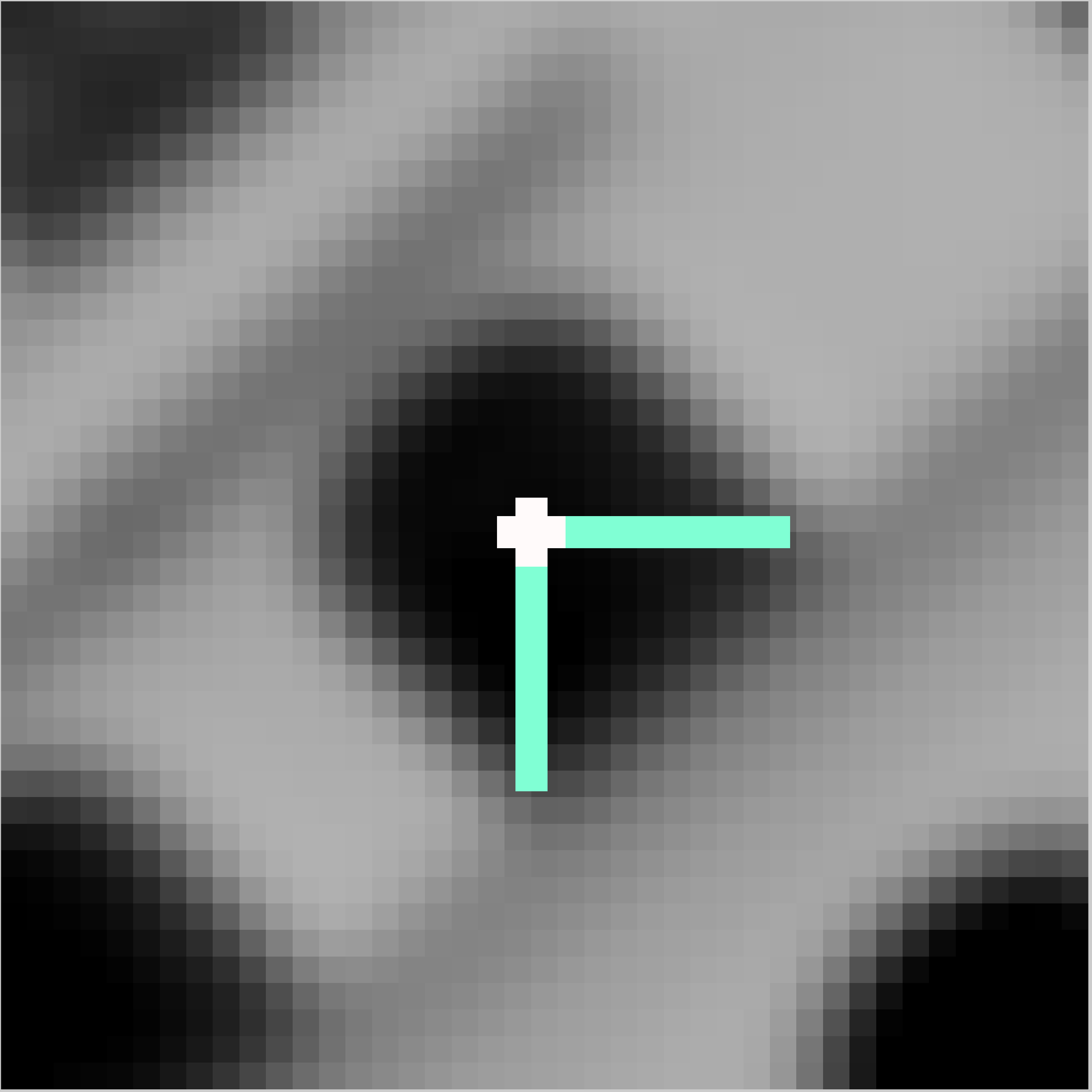}\\
} 
&
\parbox{.1\linewidth}
{
\centering
\ic{0.99}{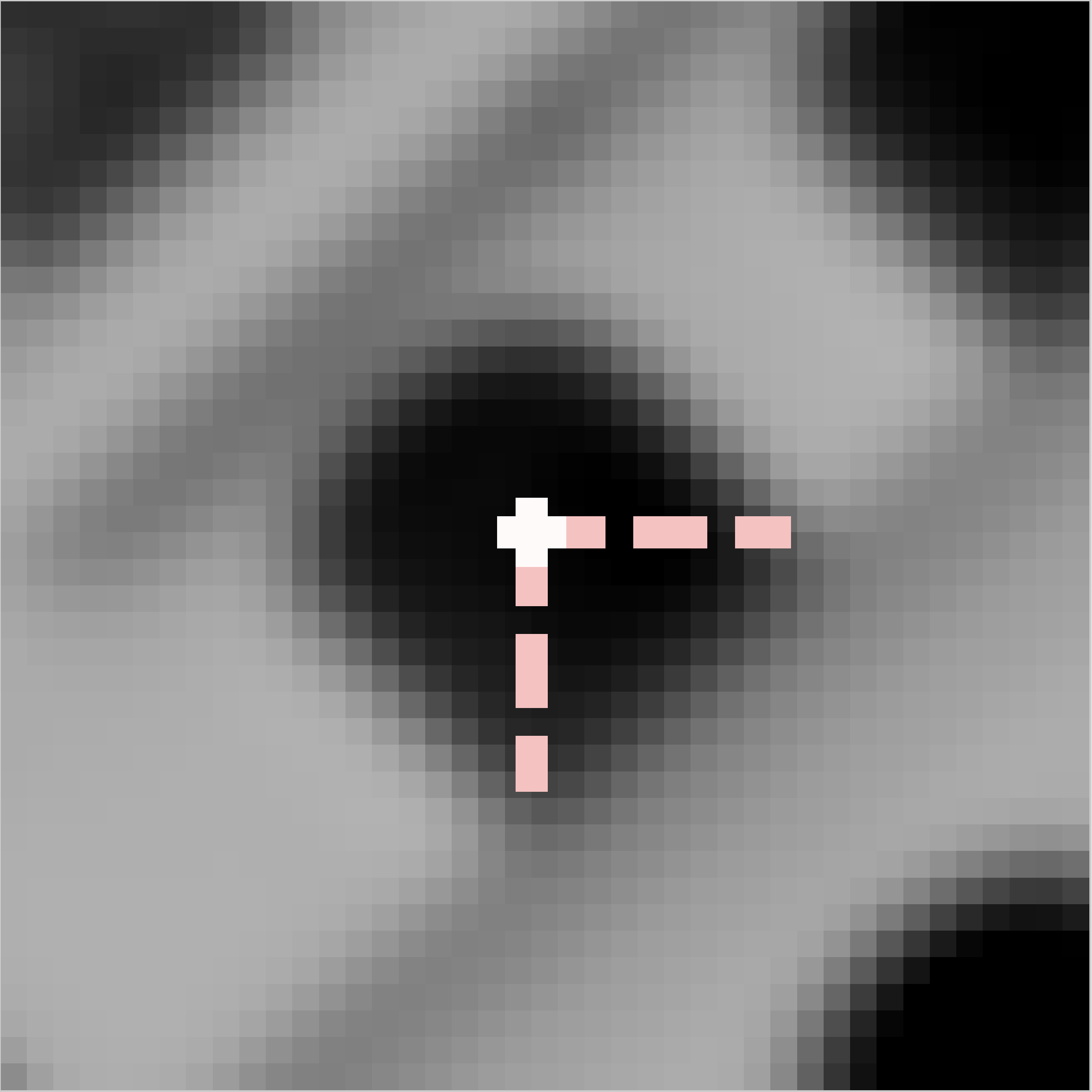}\\
\ic{0.99}{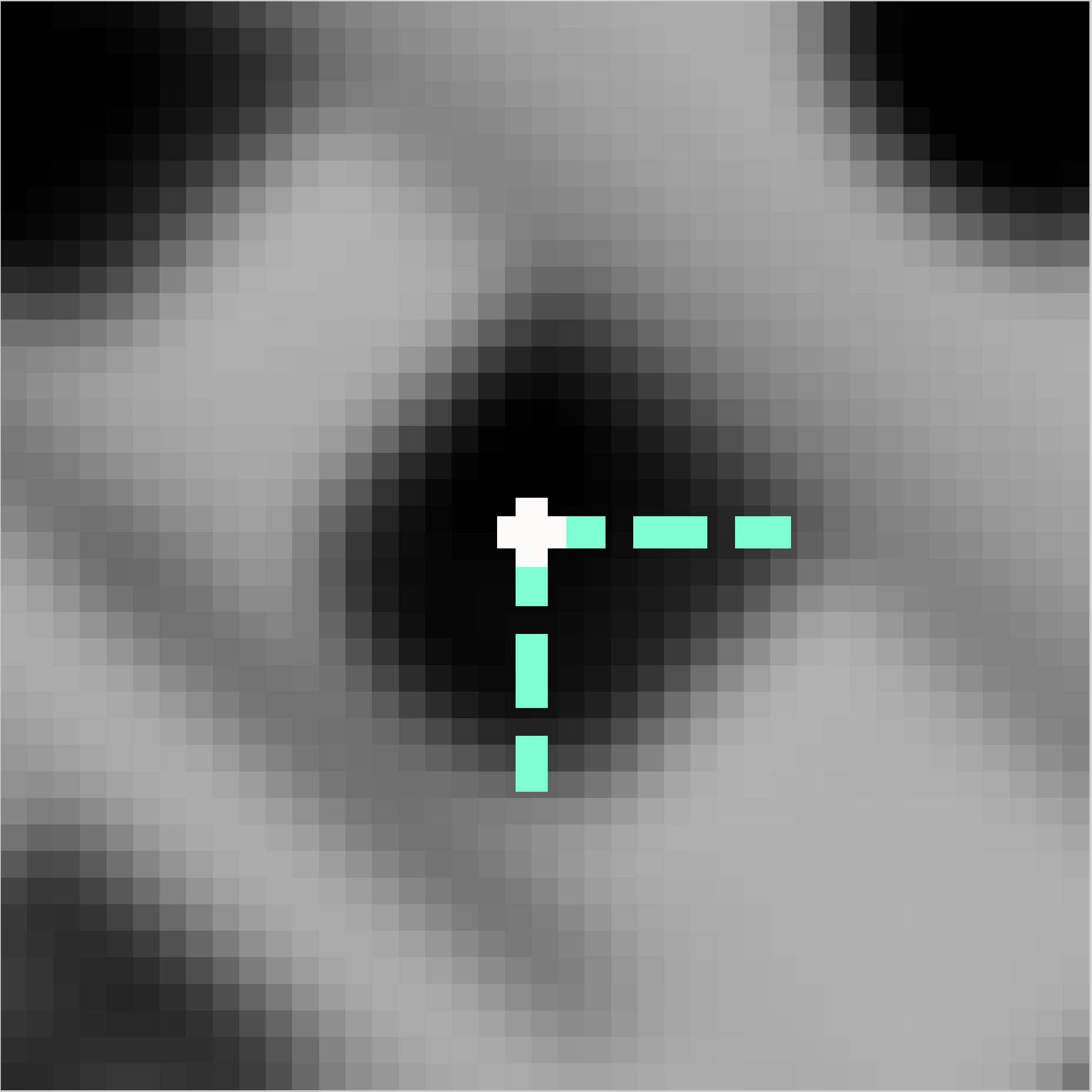}\\   
}
&
\parbox{.138\linewidth}
{\centering {\parbox{0.99\linewidth}{\includegraphics[width=\linewidth,trim={30 30 30 30}]{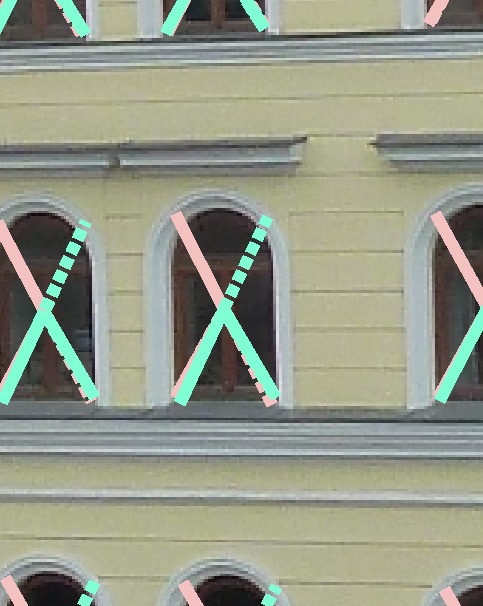}}}
}
&
\parbox{.15\linewidth}
{\centering {\parbox{0.99\linewidth}{\setlength\fboxrule{0.0pt} \fbox{\includegraphics[width=\linewidth]{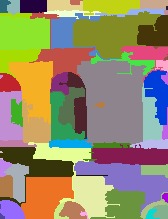}}}}
}
&
\parbox{.15\linewidth}
{\centering {\parbox{0.99\linewidth}{\setlength\fboxrule{0.0pt} \fbox{\includegraphics[width=\linewidth]{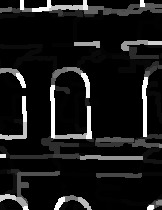}}}}
}
\\
(a) & (b) & (c) & (d) & (e) & (f) \\
\end{tabular}

\caption{Image measurements. \inlinelist{\protect\item Center of gravity (white cross) and  curvature
extrema (orange circles) of a detected MSER (orange
contour \cite{Matas-BMVC02}).  \protect\item Patches are normalized
and oriented to define an affine keypoint as in \cite{Matas-ICPR02}, and
their texture is described by
\rootsift \cite{Arandjelovic-CVPR12,Lowe-IJCV04} \protect\item Bases
are reflected for symmetry detection. \protect\item Affine keypoints
mapped back into image. \protect\item Over-segmentation by
superpixels. \protect\item The contrast feature
$\psi^{\text{contrast}}_T$, where intensity is proportional to edge
response along superpixel boundaries.}}
\end{figure}

\subsection{Measurements}
Affine-covariant
keypoints~\cite{Matas-BMVC02,Mikolajczyk-IJCV04,Obdrzalek-BMVC02} are
extracted from the image as good candidates for representing repeated
content. (see \fig 2a). The shapes of the detected patches are
summarized by keypoints, or, equivalently, 3-tuples of points, and are
given by measurements \kps. One type of keypoint construction is
illustrated in \figs 2a-d. The image is over-segmented by SEEDS
superpixels~\cite{VandenBergh-ECCV12} to provide measurements on
regions where keypoint detection is unlikely (see \fig{2e}). The
segmented regions are denoted by $\rgnls$.  The keypoints and regions
are concatenated to give the joint measurement
$\meas={\kps}^\frown \rgnls$, which is an argument to the energy
defined in \eq{\ref{eq:energy_minimization}}.

\subsection{Unary Features for Repeats and Surfaces}
\label{sec:unary_features_repeats}
The perspective skew of each scene plane is given by its vanishing
line, which is an analog to the horizon line for a scene plane at any
orientation. Vanishing lines are encoded in the parameters of the
scene planes $\beta^R(\kplblgs,\ve[y]^R)$. Explicitly they are the set
$\{\,\ve[l]_n \,\mid \, \ve[l]_n \in \mathcal{P}^2 \,\}_{n=1}^{N_V}$,
where $N_V$ is the number of scene planes and $\mathcal{P}^2$ is the
real projective plane.

\paragraph{Scale of coplanar repeats.}
A coplanar repeat group $C$ is the set of keypoints from the same pattern
that co-occur on a scene plane, namely $C
= \{\, \kp{i} \,\mid\,\kplblg{ig}=m \, \wedge \, \kplblg{iv}=n \,\}$,
where $n \neq b$. The keypoints of $C$ are called coplanar repeats. The
coplanar repeats of $C$ are of equal scale (equiareal) if their
perspective skew is removed, which is accomplished by transforming the
vanishing line of the underlying scene plane $\ve[l]_n$ so that it is
coincident with the principal axis of the camera (see
Chum \etal \cite{Chum-ACCV10}). The scale feature
$\psi^{\text{scale}}$ measures the mutual compatibility of coplanar
repeats with the scale constraint. Let $H_{\ve[l]_n}(\cdot)$ be a
transformation that removes perspective skew from plane $n$ by
orienting $\ve[l]_n$ to the principal axis and $s(\cdot)$ be the
function that computes the scale of a keypoint. Then the scale feature
for the scene's coplanar repeats is
\begin{equation}
\label{eq:keypoint_scale}
\psi^{\text{scale}} = -\sum_{m=1}^{N_G} \, \sum_{n=1}^{N_V} \, \sum_i \,  [\kplblg{ig}=m]\cdot [\kplblg{iv}=n]  \cdot \left(\log s(H_{\ve[l]_n}(\kp{i}))-\log \, \bar{s}(n,\kplblg{ig}) \right)^2,
\end{equation}
where $\bar{s}(n,\kplblg{ig})$ is the geometric mean of the keypoints in
pattern $\kplblg{ig}$ rectified by transformation
$H_{\ve[l]_n}(\cdot)$, which is part of the estimated parameters of
the repeated scene content encoded in $\beta^F(\kplblgs)$.

\paragraph{Appearance of patterns.}
The appearance of the image patches containing the keypoints \kps are
 described by \rootsift ~\cite{Arandjelovic-CVPR12,Lowe-IJCV04}. The
 corresponding \rootsift of a keypoint is given by the function
 $u(\cdot)$. The appearance affinity of keypoint $\kp{i}$ to a pattern
 is given by the normalized Euclidean distance between the \rootsift
 descriptor of the keypoint and mean \rootsift descriptor of the
 pattern. The appearance feature for patterns is
\begin{equation}
\label{eq:keypoint_texture}
{\psi}^{\text{app}} =\sum_{m=1}^{N_G} \, \sum_i \, [\kplblg{ig}=m] \cdot \frac{\|u(\kp{i})-\bar{u}(\kplblg{ig})\|_2^2}{\sigma_1^2}, \\
\end{equation}
where $\bar{u}(\kplblg{ig})$ is the mean of the \rootsifts of keypoints
in pattern $\kplblg{ig}$, which is part of the estimated parameters
of the repeated scene content encoded in $\beta^F(\kplblgs)$. The
variance $\sigma_1^2$ is set empirically.

\paragraph{Color of scene surfaces.}
The color distribution of each scene surface is modeled with a RGB
Gaussian mixture model (GMM) with $K$ components, $\gamma
= \{\, \mu_{nk},\Sigma_{nk},\pi_{nk}, \}$, where $nk \in \{\, 1 \dots
N_V, \, b \, \} \times \{\,1 \ldots K\,\}$ and
$\mu_{nk},\Sigma_{nk},\pi_{nk}$ are the mean RGB color, full color
covariance and mixing weight for component $k$ of surface $v$. The set
of GMM parameters $\gamma$ is part of the estimated parameters of the
appearance and geometry for scene planes encoded in $B^R(Y^K,Y^R)$.
The color feature for the scene surfaces is
\begin{equation}
\label{eq:color_unary}
\psi^{\text{color}} = \sum_{n \in \{1\ldots N_V,b\}} \,\sum_j \, \sum_{j'} \, \frac{[\rgnlblg{jv}=n]}{|\rgnl{j}|} \cdot \underbrace{\min_{k \in \{1,\ldots,K\}} \left\{\,-\log\left(p_n(\rgnl{jj'}|k)\cdot\pi_{nk}\right)\, \right\}}_{\text{approximately $\propto -\log p_n(\rgnl{jj'})$}},
\end{equation}
where \rgnl{jj'} is the $j'$-th member pixel of region $\rgnl{j}$ with
$|\rgnl{j}|$ number of pixels and the conditional likelihood of a
pixel \rgnl{jj'} given a mixture component $k$ is normally
distributed,
$\rgnl{jj'}|k \sim \mathcal{N}(\mu_{nk},\Sigma_{nk})$. The feature
$\psi^{\text{color}}$ uses the same approximation for the
log-likelihood as Grabcut \cite{Rother-ACM04} to make the
maximum-likelihood estimation of GMM parameters faster. Connected
components of regions with the same surface assignment segment the
image into contiguous planar and background regions.

\paragraph{Planar and background singletons.} 
Singletons are keypoints that don't repeat. A weighted cost for each
singleton is assessed, which is the maximum unary energy that can be
considered typical for a coplanar repeat. For a complete geometric
parsing of the scene, it is necessary to assign each singleton to its
underlying scene plane or to the background surface.  Singletons
induce no single-view geometric constraints nor appearance constraints
because they are not part of a repeat group, so their assignments to
scene planes are based on their interactions with neighborhood
keypoints and regions, which are defined
in \secref{\ref{sec:pairwise_features}} as assignment regularization
functions. An additional weighted cost for each planar singleton is
assessed, which is the minimum amount of required evidence obtained
through interactions with neighboring keypoints and regions to
consider a singleton planar.

\subsection{Pairwise}
\label{sec:pairwise_features}
The pairwise features are a set of bivariate Potts functions that
serve as regularizers for keypoint and region assignment to scene
model components.

\paragraph{Keypoint contrast.} 
\label{sec:keypoint_contrast}
The keypoint contrast feature penalizes models that over-segment similar
looking repeats. The keypoint contrast of the scene is
\begin{equation}
\label{eq:spatial_pwise}
\psi^{\text{contrast}}_F = \sum_{i \neq i'}[\kplblg{iv} \neq
  \kplblg{i'v}] \cdot \exp\left[-\frac{\|u(\kp{i})-u(\kp{i'})\|_2^2}{\sigma_2^2}\right],
\end{equation}
where the variance $\sigma_2^2$ is set empirically.

\paragraph{Region contrast.}
Regions have bounded area, so there may be large areas of low texture
on a scene plane or in the background that are over-segmented. Regions
that span low-texture areas can be identified by a low cumulative edge
response along their boundary. The cumulative edge response between
two regions, denoted $\phi(\rgnl{j},\rgnl{j'})$, is robustly
calculated so that short but extreme responses along the boundary do
not dominate (see \fig 2f). The region contrast of the image is given
by the feature
\begin{equation}
\label{eq:spixel_contrast}
\psi^{\text{contrast}}_R = \sum_{j \neq j'}[\rgnlblg{jv} \neq \rgnlblg{j'v}] \cdot \exp\left[-\frac{\phi(\rgnl{j},\rgnl{j'})^2}{\lambda }\right].
\end{equation}
A larger constant $\lambda$ increases the amount of smoothing and is
set as $\lambda = 2 \cdot \bar{\phi^2}$, which puts the crossover
point of smoothing at the mean contrast of regions.

\paragraph{Keypoint overlap.}
\label{sec:keypoint_overlap}
A keypoint that overlaps a region is coplanar or co-occurs on the
background surface with the overlapped region, which is encoded as a
pairwise constraint. A penalty for each violation of the coplanarity
constraint is assessed.

\subsection{Label subset costs}
\label{sec:label_subset_costs}
Parsimonious scene models are encouraged by assessing a cost for each
scene model part. Equivalence classes of the label set are defined by
labels that share a scene model part, \eg, the set of labels that have
the same vanishing line. A label subset cost is assessed if at least
one label from an equivalence class is used, which is equivalent to
accumulating a weighted count of the number of unique scene model
components in the scene.

\section{Energy Minimization}
\label{sec:energy_minimization}
The energy minimization task of \eq{\ref{eq:energy_minimization}} is
solved by alternating between finding the best labeling $\lblg$ and
regressing the scene model components $\beta$ in a block-coordinate
descent loop until the energy converges. Alternating between
finding the minimal energy labeling and regressing continuous model
parameters has notably been used in segmentation and multi-model
geometry estimation by Rother \etal and Isack \etal
\cite{Rother-ACM04,Isack-IJCV12}.

\subsection{Labeling and Regression}
The scene model parameters are fixed to the current estimate for the
labeling problem, $\hat{\ve[y]}=\argmin_{\ve[y]}
\,E(\meas,\lblg,\beta(\ve[y])=\hat{\beta})$. Finding the
minimal-energy labeling is NP-hard \cite{Boykov-PAMI01a}. An extension
to alpha-expansion by Delong \etal
\cite{Boykov-PAMI01a,Delong-IJCV12,Kolmogorov-PAMI04} that
accommodates label subset costs (defined in
\cref{sec:label_subset_costs}) is used to find an approximate
solution.

The labeling is fixed to the current estimate for the regression
subtask
$\hat{\beta}=\argmin_{\beta}\,E(\meas,\lblg=\hat{\lblg},\beta(\hat{\lblg})).$
Each continuous parametric model must be regressed with respect to its
dependent unary potentials so that the energy does not increase during
a descent iteration. In particular, the vanishing lines, surface color
distributions and the representative appearance for patterns and
rectified scale for coplanar repeats are updated as detailed in the
following paragraphs. The updated parameters are aggregated in
$\hat{\beta}$.

\paragraph{Vanishing lines.} 
All keypoints assigned to the same planar surface are used to refine the
surface's vanishing line orientation. The objective is the same as the
unary defined in eq.~\ref{eq:keypoint_scale} and encodes the affine scale
invariant defined in Chum \etal \cite{Chum-ACCV10}. The vanishing line
is constrained to the unit sphere and so that all keypoints are on the
same side of the oriented vanishing line,
\begin{gather}
  \label{eq:refine_vlines}
  \ve[l]^*_n = \argmin_{\ve[l]} \sum_{i\,:\,\hkplblg{iv}=n} \left( \log s(H_{\ve[l]}(\kp{i}))
  -\frac{1}{\sum_{i'}\,[\hkplblg{ig}=\hkplblg{i'g}]} \log \sum_{i'}\,[\hkplblg{ig}=\hkplblg{i'g}] \cdot s(H_{\ve[l]}(\kp{i'}))
  \right)^2 \\
  \begin{aligned}
    \textup{s.t.}  \quad \ve[l]^{\T}\kp{iw} &> 0, \quad  w \in \{\, 1 \dots 3 \, \} \\ 
    \ve[l]^{\T}\ve[l] &= 1, \\
  \end{aligned}
\end{gather}
for all scene planes $n$ that have patterns assigned, where $s(\cdot)$
is the scale of a keypoint and $H_{\ve[l]}(\cdot)$ is the rectifying
transform as defined in \cref{sec:unary_features_repeats}, and
\kp{iw} denotes the individual homogeneous coordinates that define
keypoint \kp{i}. The constrained nonlinear program is solved with the
MATLAB intrinsic \textsc{fmincon}.

\paragraph{Coplanar repeats and patterns.}
For features $\psi^{\text{scale}}$ eq.~(\ref{eq:keypoint_scale}) and
$\psi^{\text{app}}$ eq.~(\ref{eq:keypoint_texture}) that are sums of
squared differences, the parameters are estimated as a mean of the
respective values. 

\paragraph{Surface color distribution.}
The parameters of the color distribution of a surface are estimated
from the member pixels of regions assigned to the surface. The
approximate log-likelihood defined for the unary $\psi^{\text{color}}$
in eq. $\ref{eq:color_unary}$ is maximized to estimate the Gaussian
mixture for each surface that has region assignments,
\begin{equation}
\label{eq:regress_color}
  \{\,\Sigma^*_{nk},\mu^*_{nk},\pi^*_{nk}\,\}_{k=1}^K = \argmax_{\{\Sigma_{nk},\mu_{nk},\pi_{nk}\}_{k=1}^K} \, \prod_{j : \hrgnlblg{j} = n} \, \prod_{j'}\max_{k'} \, p_v(\rgnl{jj'} \mid k';\Sigma_{nk},\mu_{nk},\pi_{nk}) \cdot \pi_{nk'}.
\end{equation}
The objective defined in eq.~\ref{eq:regress_color} is maximized by
block-coordinate ascent in a manner similar to Lloyd's algorithm: The
mixture component assignments are fixed to estimate the means and
covariances and then vice-versa in alternating steps. A fixed number
of iterations is performed.

\subsection{Proposals}
The initial minimal labeling energy requires a guess $\beta^{{0}}$ at
the continuous parameters $\beta(\lblg)$. This is provided by a
proposal stage in which the keypoints \kps are clustered by their
\rootsift descriptors and sampled to generate vanishing line
hypotheses as in Chum \etal \cite{Chum-ACCV10}. The clustered regions
are verified against the hypothesized vanishing lines to create a
putative collection of coplanar repeats that are scale-consistent
after affine rectification by a compatible sampled vanishing line. The
proposed coplanar repeat groups do not partition the keypoints, which
is a constraint enforced by the minimal energy labeling
$\ve[\hat{y}]$. The inital color model for each detected surface
(equivalently proposed vanishing lines and background) is estimated
from the image patches of keypoints from the proposed coplanar repeat
groups.

\section{Dataset}
We introduce a dataset\footnote{Available
at \href{http://ptak.felk.cvut.cz/personal/prittjam/bmvc16/coplanar\_repeats.tar.gz}{http://ptak.felk.cvut.cz/personal/prittjam/bmvc16/coplanar\_repeats.tar.gz}
} of 113 images containing from 1 to 5 scene planes with translated,
reflected and rotated coplanar repeats occurring periodically or
arbitrarily. The dataset includes some images from the ZuBuD database
of Shao \etal and the CVPR 2013 symmetry database assembled by 
Liu \etal \cite{shao2003zubud, liu2013symmetry, park2010translation}.
The manual assignment of keypoints to coplanar repeat groups is
infeasible since a typical image will have thousands of extracted
keypoints. Direct annotation is also undesirable since setting changes
of the keypoint detectors would invalidate the assignments. Instead,
the annotations are designed to constrain the search for coplanar
repeated keypoints, making annotations agnostic to the keypoint
type. The annotations hierarchically group parallel scene planes,
individual scene planes, and areas within a scene plane that cannot
mutually have the same coplanar repeats, \ie denoting distinct
patterns. Clutter and non-planar surfaces are also
segmented. Keypoint-level assignment to coplanar repeat groups is
achieved using a RANSAC-based estimation framework which leverages the
annotations to constrain the search for correspondences to choose the
correct transformation type.

\begin{figure}[t!]
\label{fig:img1_ann}
\begin{tabular}{c@{}c@{}c@{}c@{}c@{}}
 \ic{0.2}{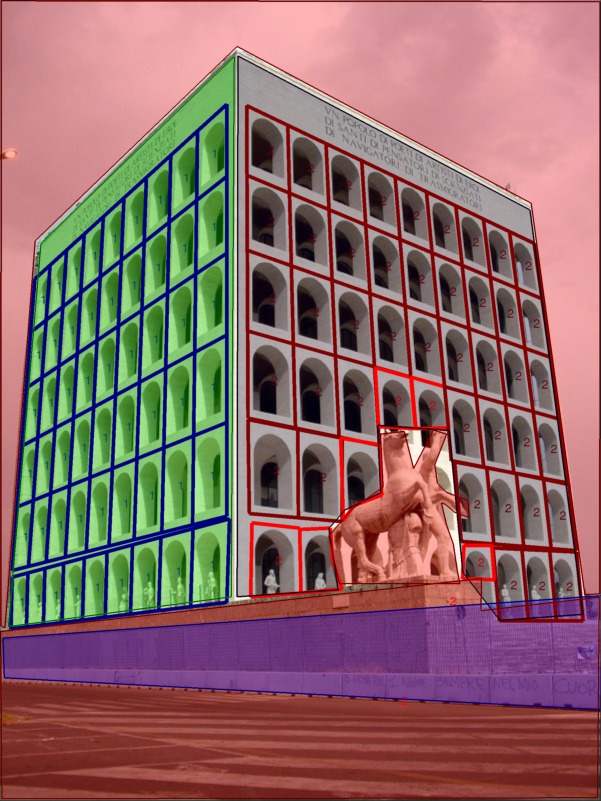} &
 \ic{0.2}{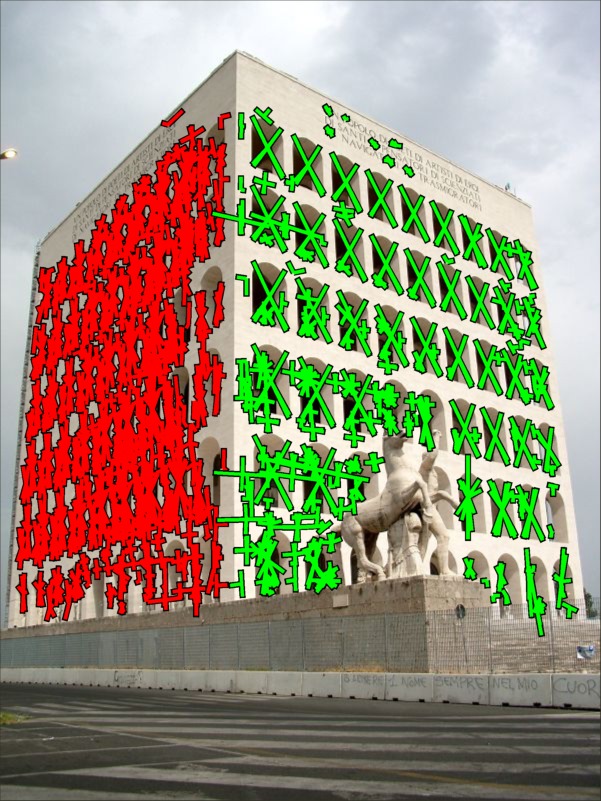} &
 \ic{0.2}{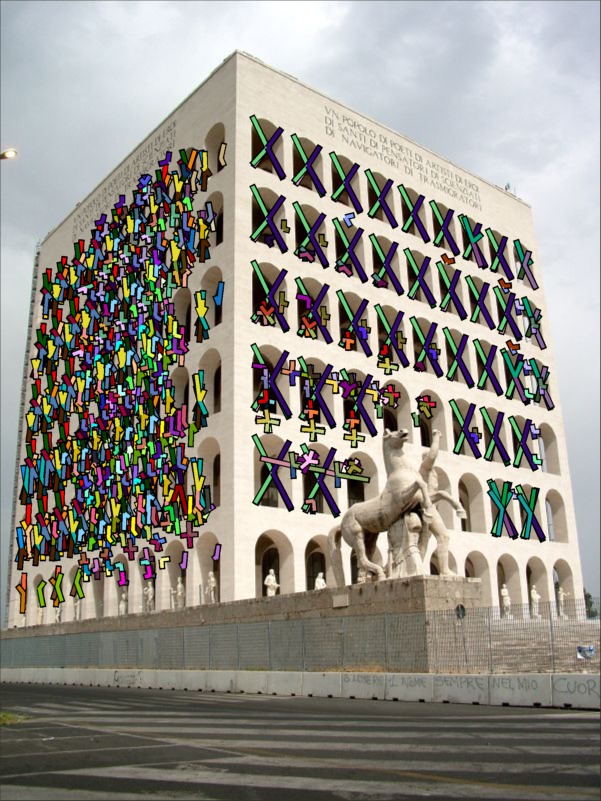} &
 \ic{0.2}{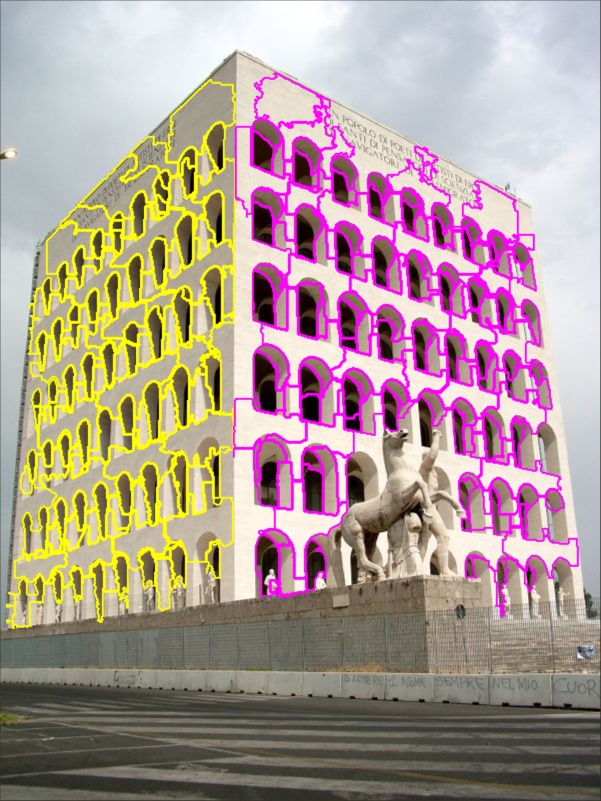} &
 \ic{0.2}{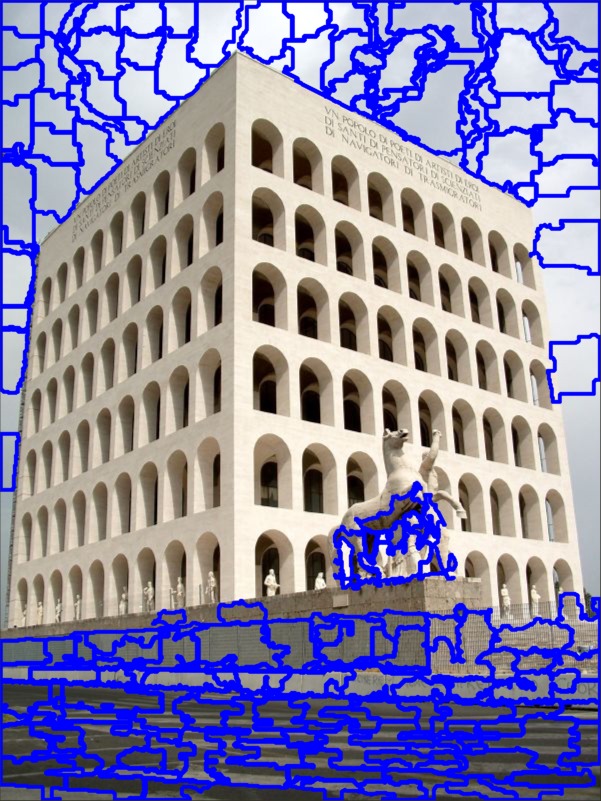} \\
 (a) & (b) & (c) & (d) & (e) \\
\end{tabular}
\caption{The hierarchical annotations included with the 113 image dataset. 
 (a) translation symmetries are annotated by grids, regions that
 cannot share coplanar repeats are colored differently,  
 (b) detected keypoints to vanishing line assignment,
 (c) groups of coplanar repeated keypoints found by annotation-assisted inference,
 (d) image regions (SEEDS superpixels~\cite{VandenBergh-ECCV12}) to vanishing line assignment, 
 (e) and background image regions, which coplanar repeats cannot overlap.}
\end{figure}

\section{Evaluation}
\label{sec:Evaluation}
We evaluate the proposed method against two state-of-the-art geometric
multi-model fitting methods: J-Linkage and
MultiRANSAC \cite{Zuliani-ICIP05,Toldo-ECCV08}.  Both estimators are
hypothesize-and-verify variants. A model hypothesis consists of a
vanishing line and tentatively grouped keypoints of similar
appearance. Coplanar repeat group assignments are verified by a
threshold test on the similarity measure for repeated keypoint
detection proposed by Shi \etal \cite{Shi-CVPR15}. However, the
rectified scale constraint defined in Eq. \ref{eq:keypoint_scale} is
used in lieu of the scale kernel used by \cite{Shi-CVPR15}.  We
provide the number of scene planes present in each image to
MultiRANSAC.

The accuracy of rectifications constructed from vanishing lines
computed from detected coplanar repeat groups are used to compare the
methods. Two necessary conditions for accurate rectifications are
that\begin{inparaenum}[(i)]\item no outliers are included in the
detected coplanar repeat groups, \item and detected coplanar repeat
groups densely cover the extents of the scene plane where there are
coplanar repeat groups annotated in the dataset\end{inparaenum}. Thus
the rectification accuracy of coplanar repeats serves as a proxy
measure for the precision and recall of coplanar repeat detection.

Projective distortion is added by rewarping a set of annotated
coplanar repeats rectified by the transform computed from detected
coplanar repeat groups $\hat{H}(\cdot)$ with the inverse rectification
$H^{\inv}(\cdot)$ computed from the annotated repeats.  The amount of
distortion is measured as the square pointwise distance between the
annotated coplanar repeats and the rewarped coplanar repeats,
\begin{equation}
        \Delta^{\hat{H}}_\text{rms}
        = \sqrt{\frac{1}{3\cdot|\mathcal{A}|} \sum_{i \in \mathcal{A}} \sum_{w=1}^3
        d^2(\kp{iw},H^{\inv}(A^{\inv}(\hat{H}(\kp{iw})))},
\end{equation}
where $\mathcal{A}$ is the set of keypoint indices of the annotated
coplanar repeats used to compute $H$, $A(\cdot)$ resolves the affine
ambiguity between the original and rewarped annotated coplanar
repeats, and $d(\cdot,\cdot)$ gives the euclidean distance between
points. The set of annotated coplanar repeats that is the largest
proportion of the detected coplanar repeats is used to match the
rectification computed from detected coplanar repeats to a
rectification computed from annotated coplanar repeats.

\begin{figure}[t!]
\tribox{\includegraphics[width=\linewidth]{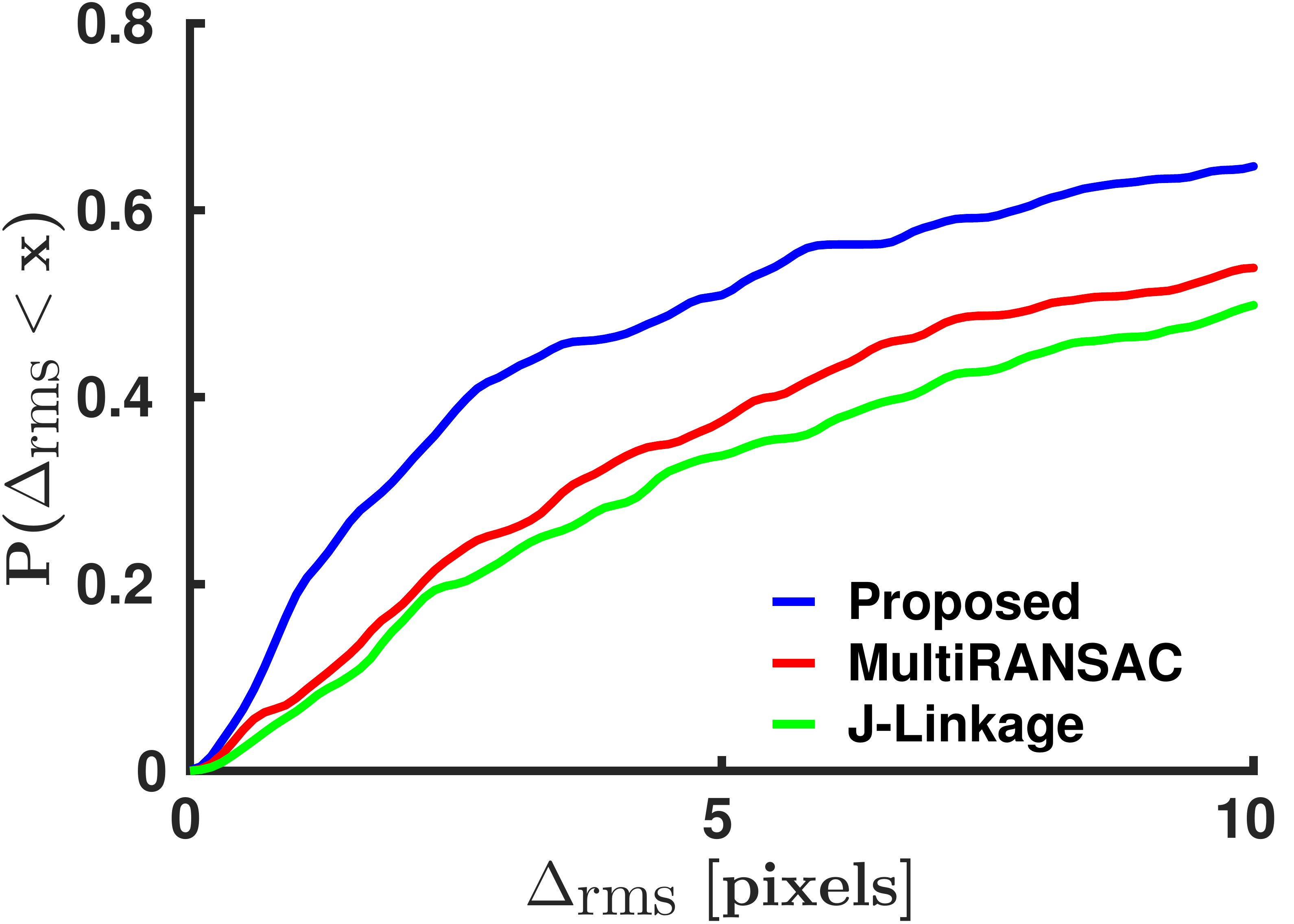}\\(a)}{\includegraphics[width=\linewidth]{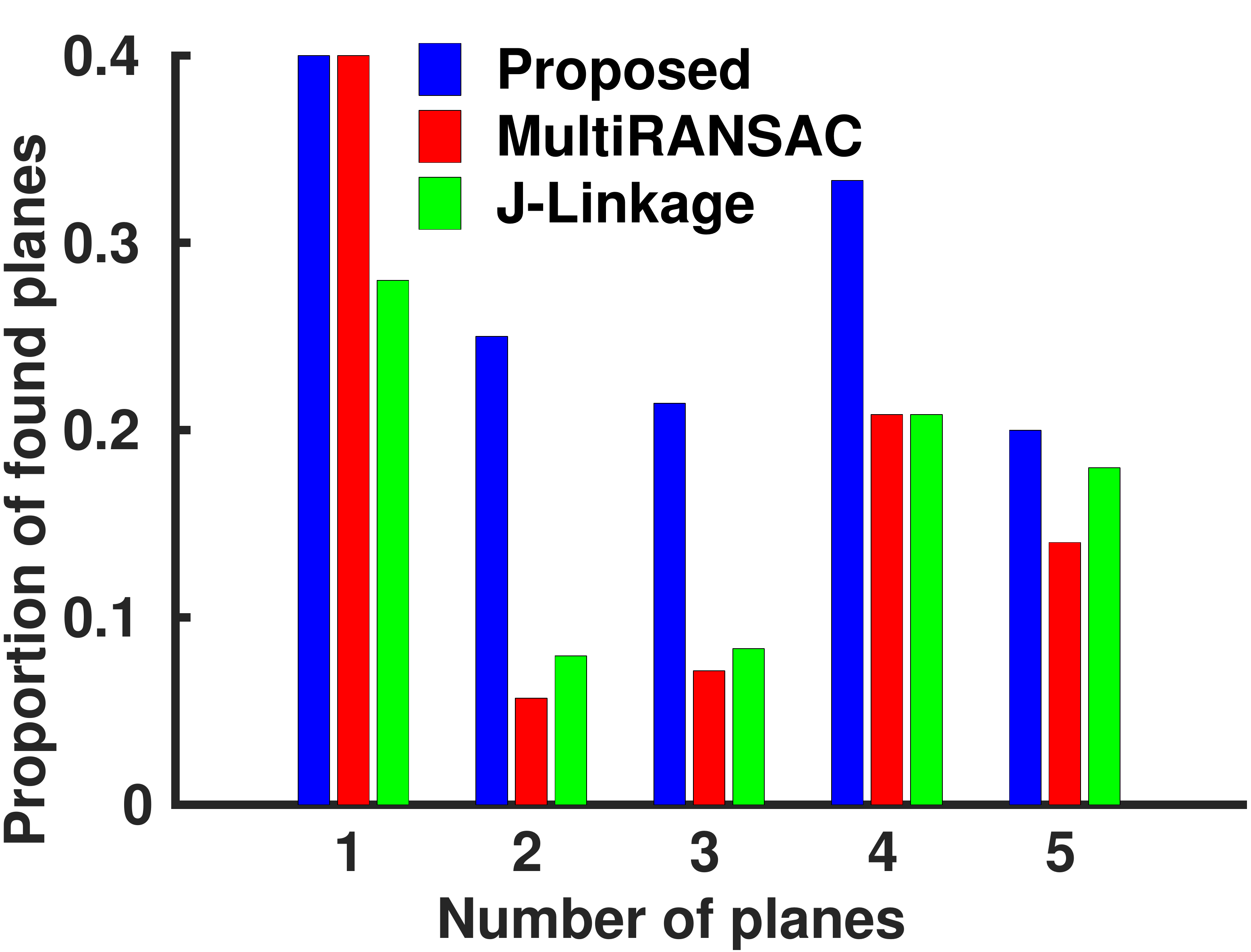}\\(b)}{\includegraphics[width=1.3\linewidth]{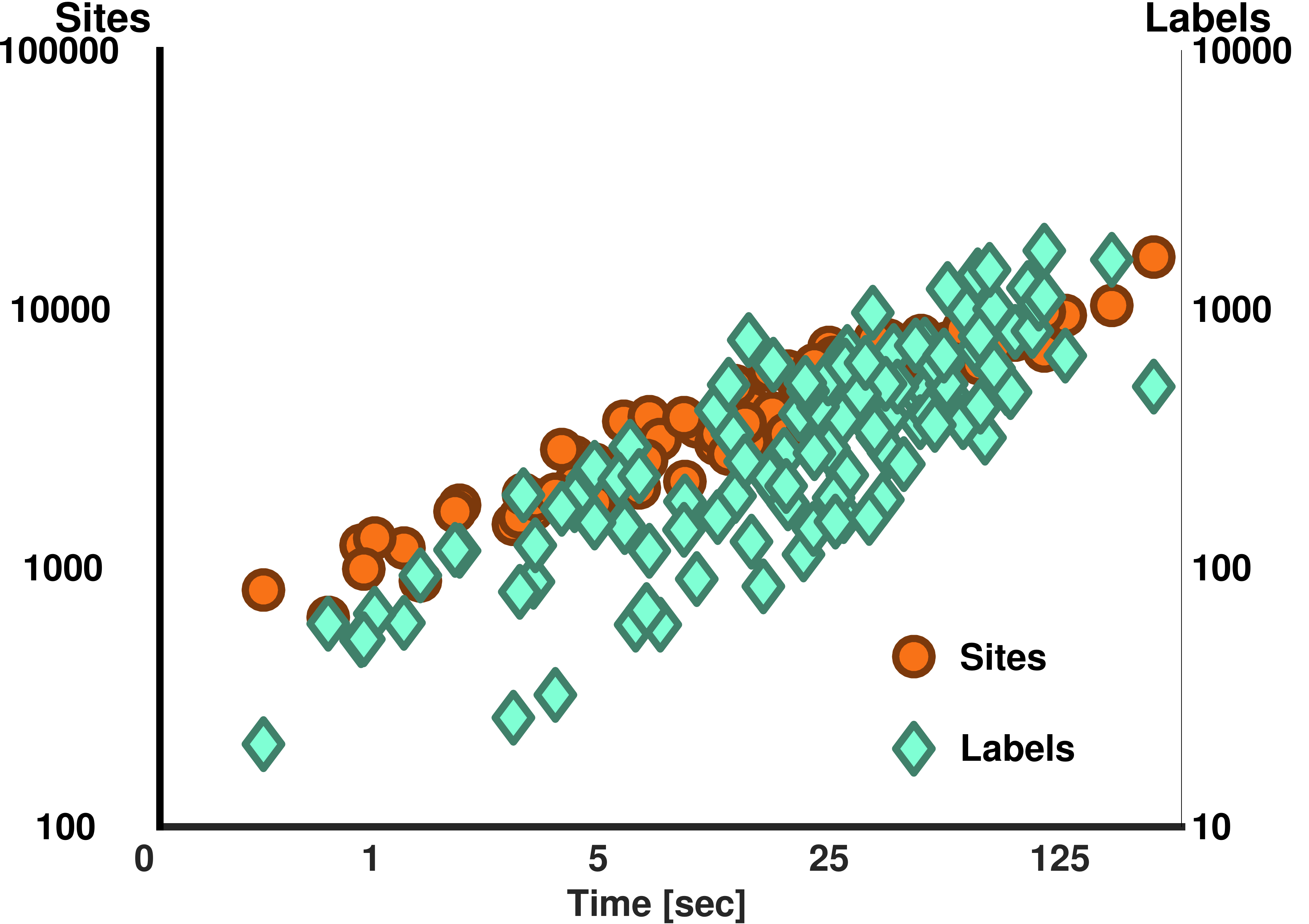}\\(c)}
\caption{Evaluation.\inlinelist{\protect\item CDF of rectification distortions ($\Delta_\text{rms}$), \protect\item proportion of planes rectified with less than 2 pixels of distortion in images with 1 to 5 scene planes, \protect\item cumulative wall time in seconds for the labeling task of energy minimization.}}
\end{figure}

The cumulative distribution of distortions on the dataset (truncated
at 10 pixels) is shown in \fig 4a. At 1 pixel of distortion, the
proposed method solves 163\% more scene planes than the next best; at
2 pixels, 94\% more; and at 5 pixels, which can be considered a
threshold for meaningful rectification, 51\% more scene planes. \fig
4b plots the proportion of scene planes rectified with less than 2
pixels of distortion with respect to the number of scene planes in the
image. Clearly the proposed method excels when there are multiple
scene planes present.  \fig 4c plots the cumulative runtime of the
labeling step for images as function of the number of keypoints and
image regions, denoted \emph{sites}, and the number of active model
proposals, denoted \emph{labels}. Inference ranges from under a second
to 2 minutes for the largest problems in the dataset.

\
\blfootnote{\textbf{Acknowledgements:} J. Pritts, D. Rozumnyi and O. Chum were supported by the MSMT LL1303 ERC-CZ.}

\section{Discussion}
The proposed energy minimization formulation demonstrates a distinct
increase in the quality of rectifications estimated from detected
coplanar repeat groups on the evaluated dataset with respect to two
state-of-the-art geometric multi-model fitting methods. The advantage
can be attributed to the global scene context that is incorporated
into the energy functional of the proposed method. The evaluation was
performed on a new annotated dataset of images with coplanar repeats
in diverse arrangements. The dataset is publicly available.

Despite a significant improvement over the baseline, the proposed
method failed to solve roughly half of the dataset with less than 5
pixels of distortion.  Future work will incorporate constraints
specific to reflected and rotated keypoints and parallel scene lines,
which would add significant geometric discrimination to the
model. Learning the feature weight vector \ve[w], which was hand
tuned, could also give a significant performance boost. However, the
complete annotation of coplanar repeated keypoints in an image is
probably infeasible. This means structured output learning must be
performed with partial annotations, which complicates the learning
task considerably.

\pagebreak

\bibliography{bmvc16.bbl}

\appendix
\numberwithin{figure}{section}
\newpage
\addtocontents{toc}{\protect\setcounter{tocdepth}{2}}
\pagestyle{plain}

\setcounter{figure}{0}
\setcounter{table}{0}
%
\onecolumn{%
\centering
\LARGE \mytitle \\ 
(Supplementary Material) \\[1.5em]
\normalsize
}

\section{Annotation-Assisted Repeat Grouping}
The annotations provided by the 113 image dataset referenced in the
paper are discussed in detail (available
at \hyperlink{http://ptak.felk.cvut.cz/personal/prittjam/bmvc16/coplanar\_repeats.tar.gz}{http://ptak.felk.cvut.cz/personal/prittjam/bmvc16/coplanar\_repeats.tar.gz}). The annotations hierarchically segment
the image into parts that \begin{inparaenum}[(i)]\item are scene
  planes \item are the union of scene planes that share the same
  vanishing line \item contain repeated content \item are the union of
  repeated content annotations that are distinctly different from
  other repeated content in the remainder of the
  image\end{inparaenum}.  In particular the repeated content
  annotations are specific to the type of symmetry exhibited by the
  repeat: namely annotations for translational and rotational
  symmetries are provided.  In addition lattices are provided for
  translationally symmetric periodic repeats.
  
  Individual salient features (\eg Hessian Affine Keypoints or MSERs) are
  not grouped or annotated, so the annotations are feature agnostic,
  which is preferable since settings adjustments would invalidate such
  annotations. Rather, the annotations are used to assist a
  RANSAC-based inference algorithm to establish coplanar repeat
  groups. The annotations constrain the search for correspondences,
  which gives a much higher inlier percentage among tentative
  groupings that are inputted to RANSAC. Since the transform type is
  known from the annotations, the transform with the fewest required
  constraints can be used, which improves the probability of proposing
  a transform estimated from all-inliers. The vanishing line is
  estimated, and, depending on the annotation tag, either a
  translation or rotation and translation, which maps repeats onto
  each pointwise.  The annotations are tagged so that the correct
  transformation can be estimated during annotation-assisted
  inference.
 
  Even with this relaxed standard of annotation, it is impossible to
  group repeats at their highest frequency of recurrence. Depending on
  the features extracted, \eg, corners of facade ornamentation may be
  detected, where only the windows were marked as repeated. Thus any
  performance evaluation must not penalize methods that correctly
  identify repeats that recur at higher frequencies than the
  annotations. Reflections and rotational symmetries, in particular,
  exacerbate this problem. Perhaps the most common example in the
  dataset are window panes, which have axial symmetry, and if square,
  rotational symmetry. It is not practical to annotate all such
  occurrences (not just restricted to windows) in a large dataset.
    The annotations also group oversegmetnations of the image (\ie
  superpixels in this context) into contiguous components of planes,
  sets of parallel planes and background surface. These annotations
  are not currently used in the evaluation, but would be useful for
  learning the regularization weights in the energy function.
\begin{figure}[H]

\begin{tabular}{c@{\hspace{3pt}}c@{}c@{}c@{}c@{}c}
	(a) &
	\ic{0.12}{img/img1_ann.jpg} &
	\ic{0.215}{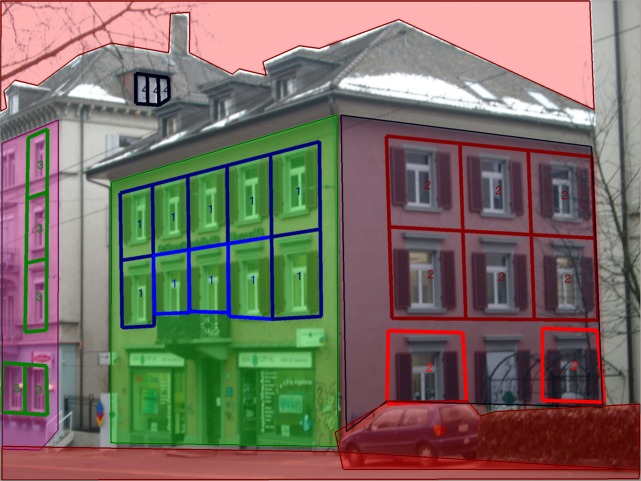} &
	\ic{0.09}{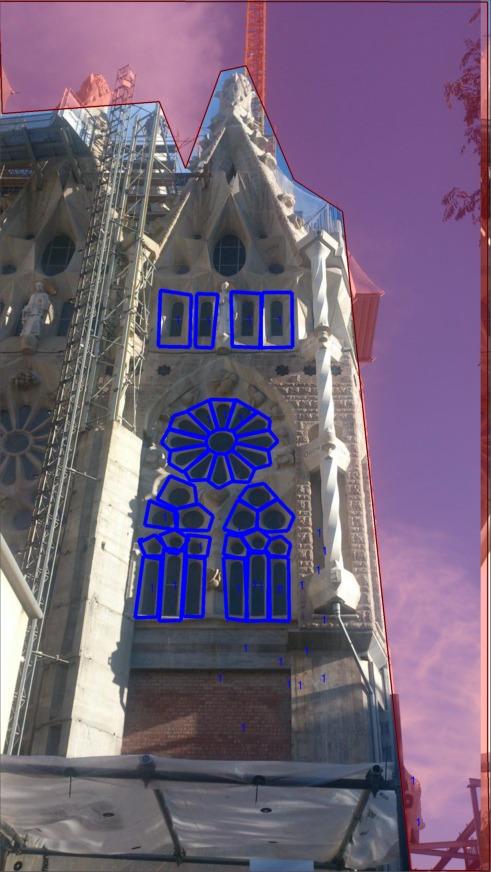} &
	\ic{0.215}{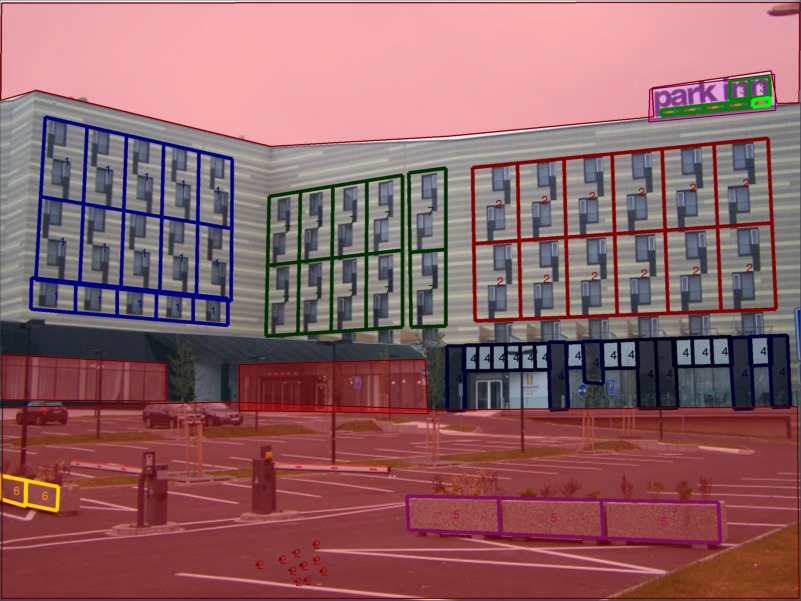} &
	\ic{0.243}{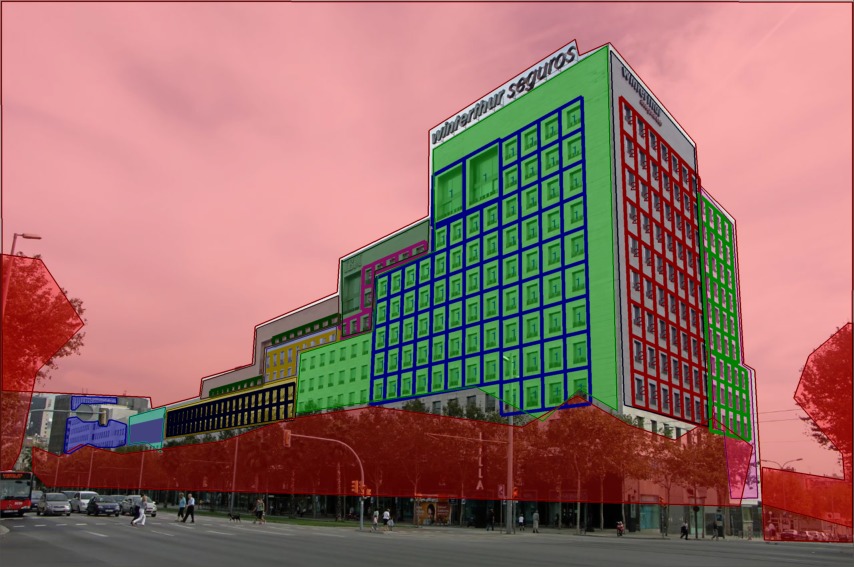} \\

    (b) &
	\ic{0.12}{img/img1_linf.jpg} &
	\ic{0.215}{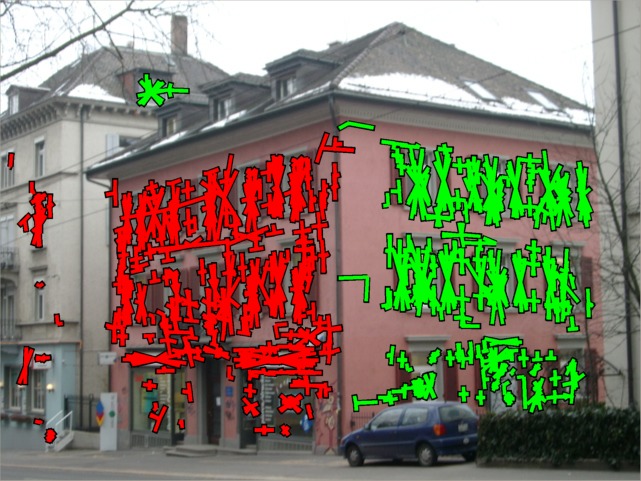} &
	\ic{0.09}{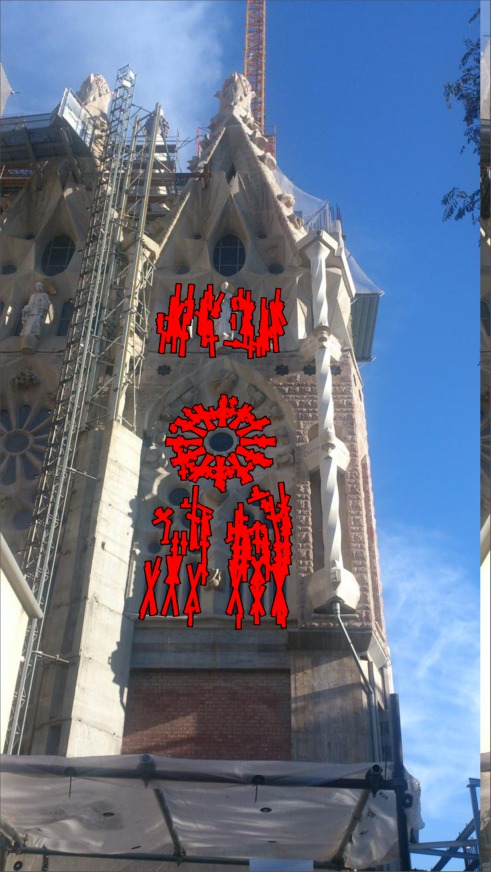} &
	\ic{0.215}{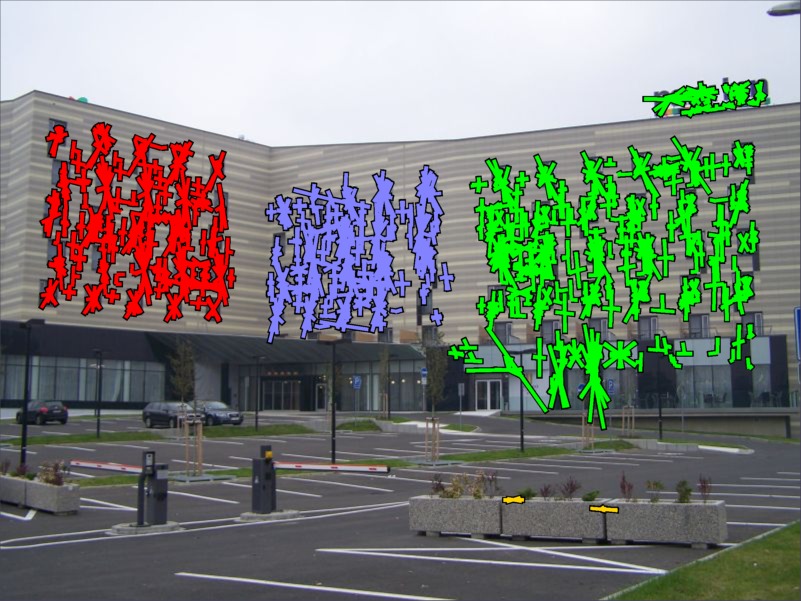} &
	\ic{0.243}{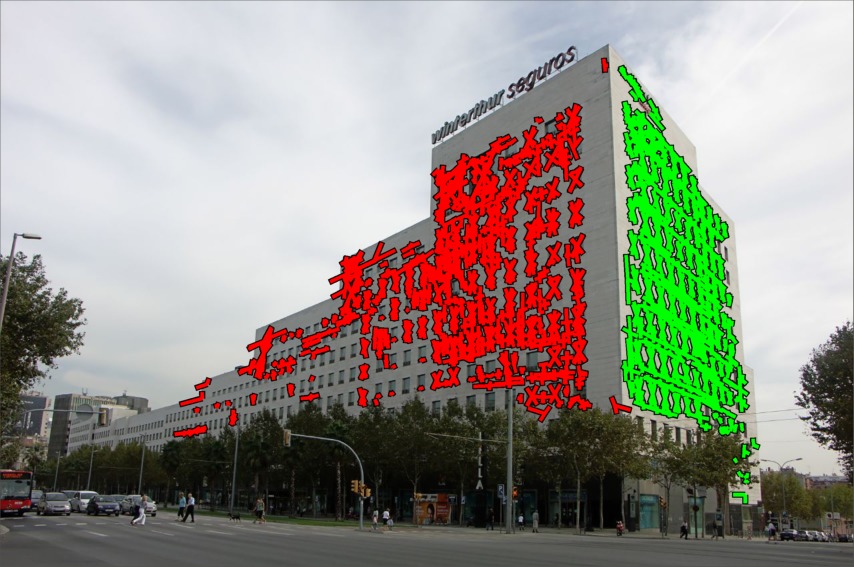} \\

    (c) &
	\ic{0.12}{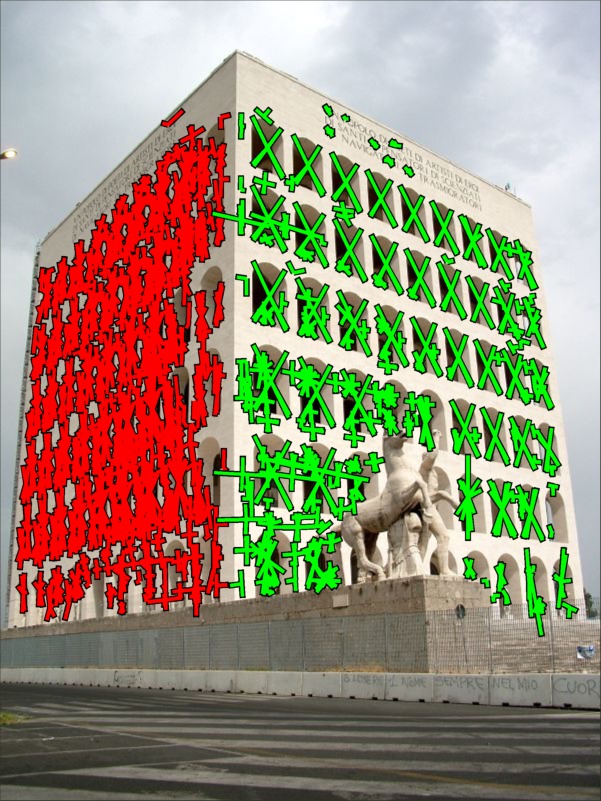} &
	\ic{0.215}{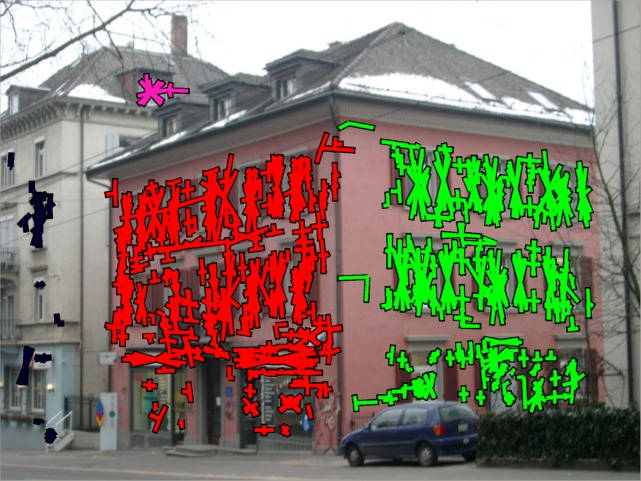} &
	\ic{0.09}{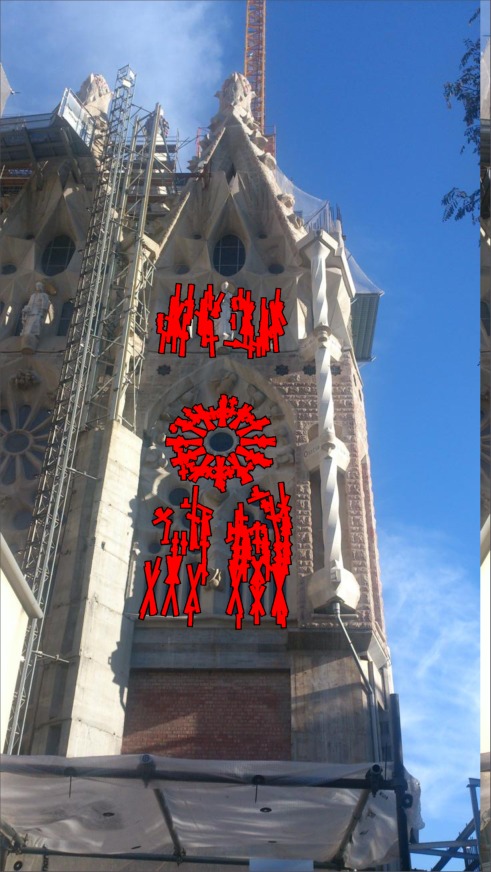} &
	\ic{0.215}{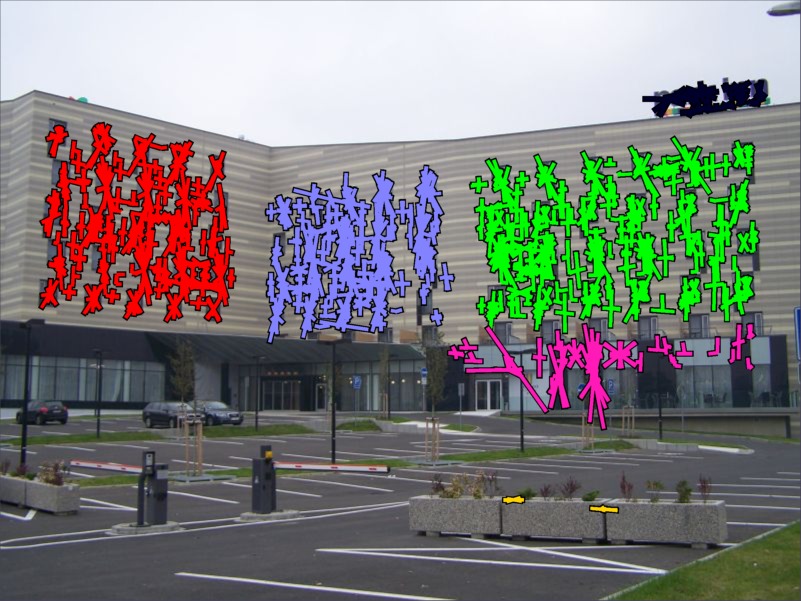} &
	\ic{0.243}{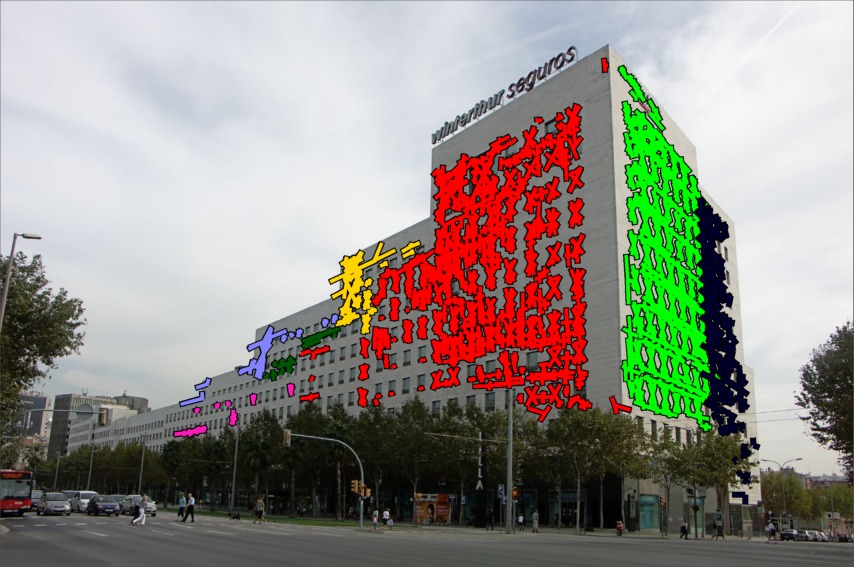} \\

    (d) &
	\ic{0.12}{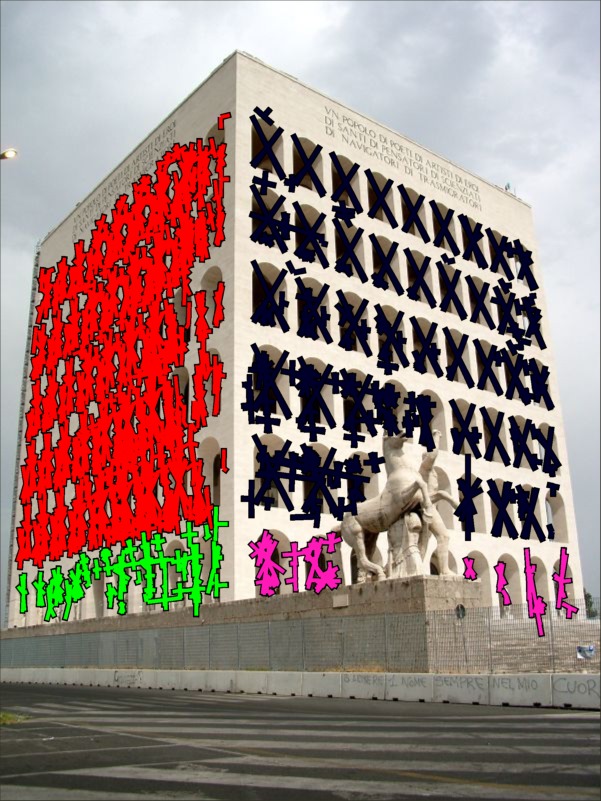} &
	\ic{0.215}{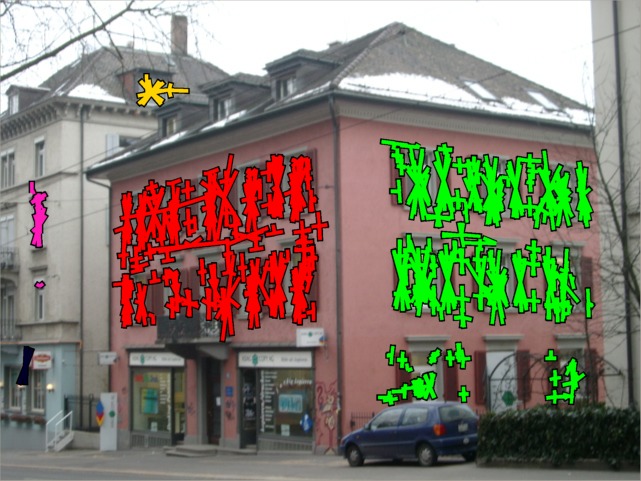} &
	\ic{0.09}{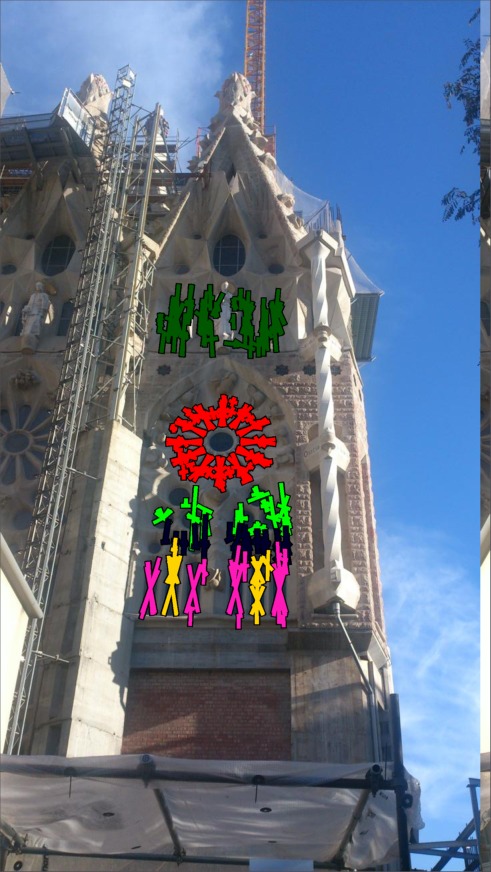} &
	\ic{0.215}{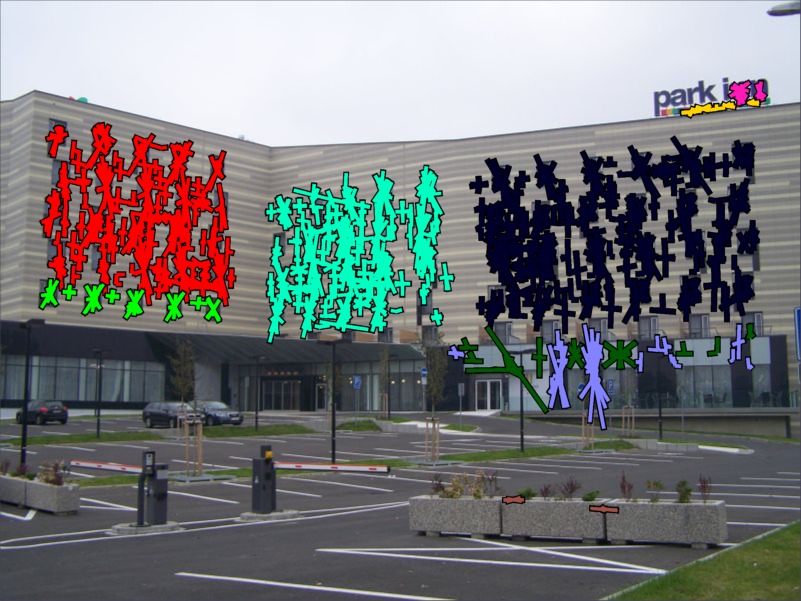} &
	\ic{0.243}{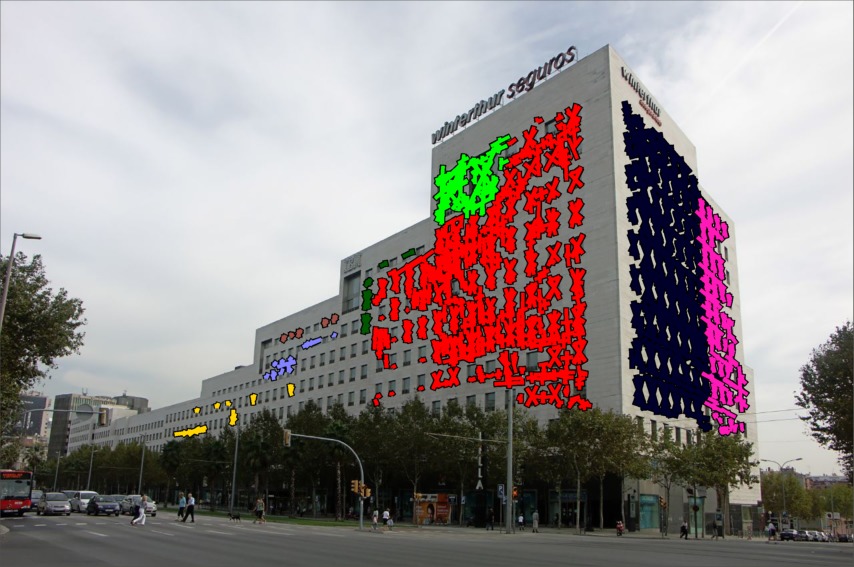} \\

    (e) &
	\ic{0.12}{img/img1_repeat.jpg} &
	\ic{0.215}{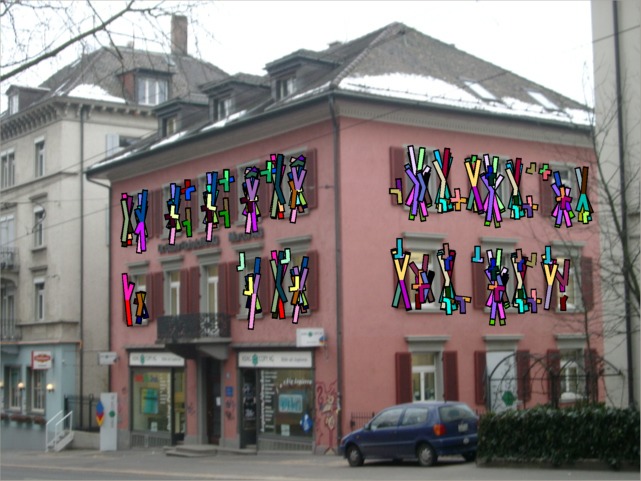} &
	\ic{0.09}{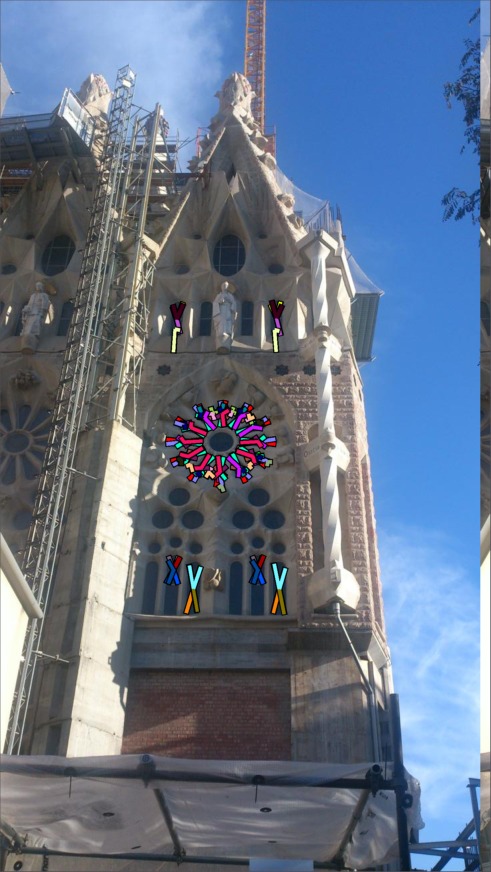} &
	\ic{0.215}{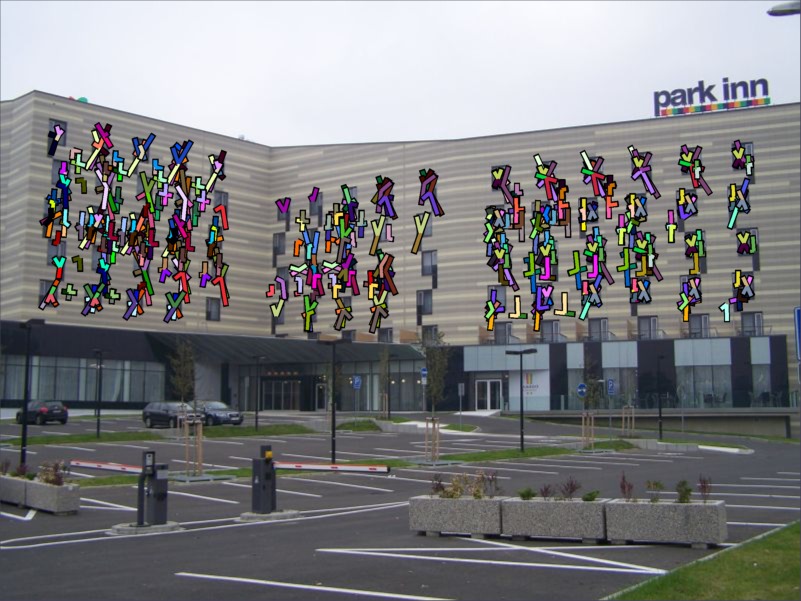} &
	\ic{0.243}{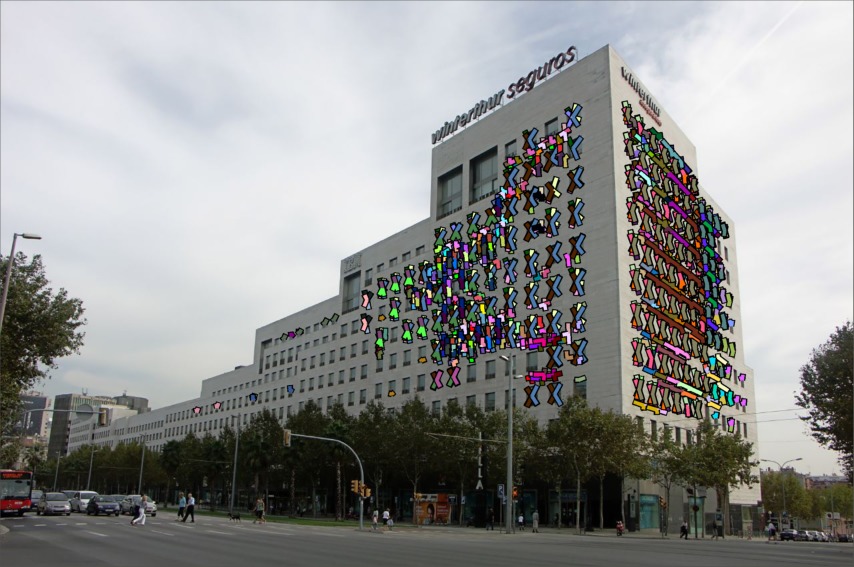} \\

    (f) &
	\ic{0.12}{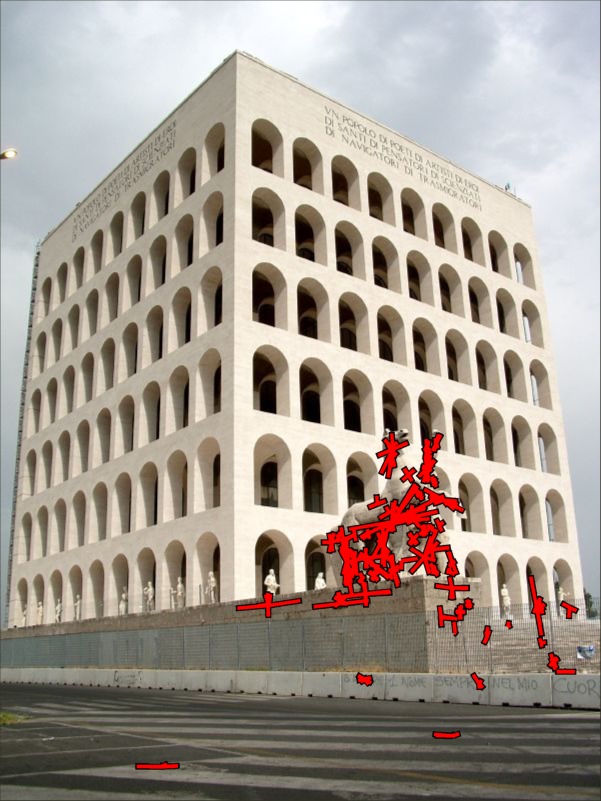} &
	\ic{0.215}{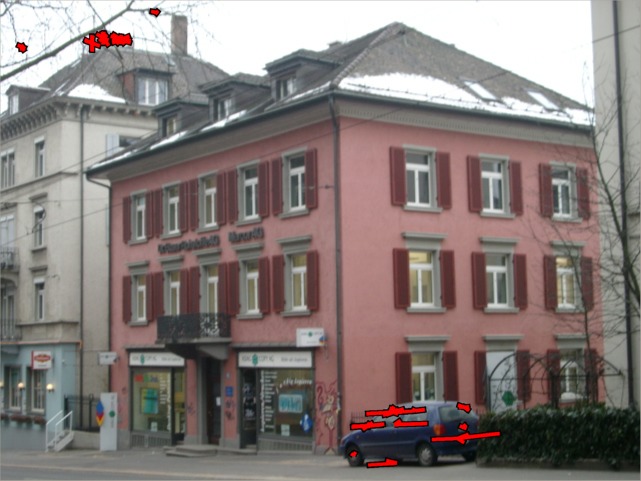} &
	\ic{0.09}{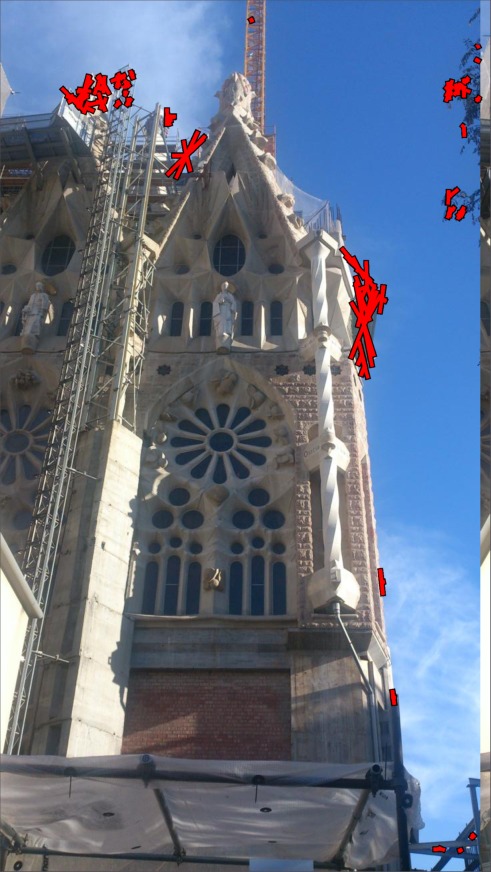} &
	\ic{0.215}{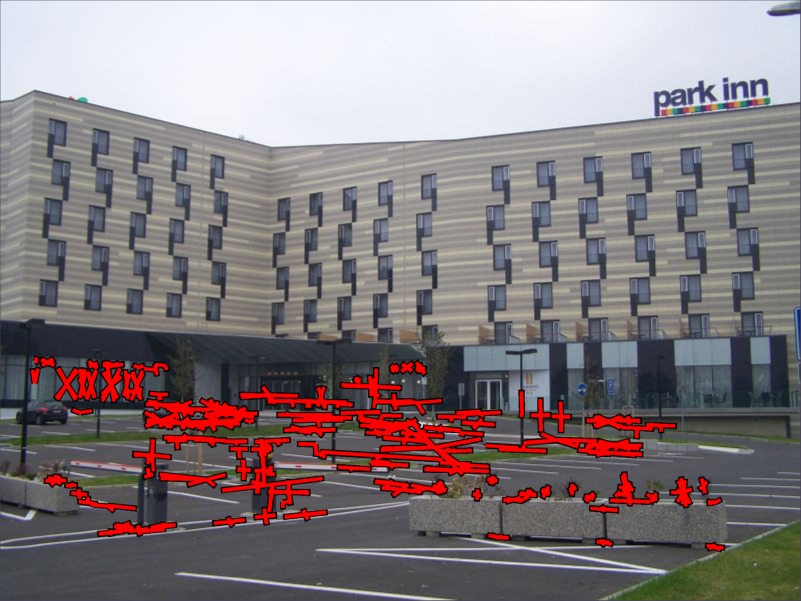} &
	\ic{0.243}{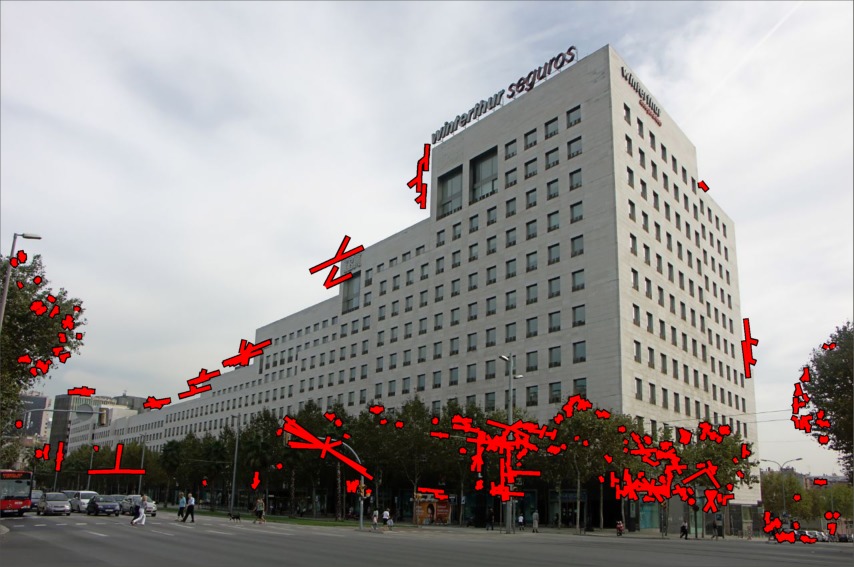} \\

    (g) &
	\ic{0.12}{img/img1_spixel_linf.jpg} &
	\ic{0.215}{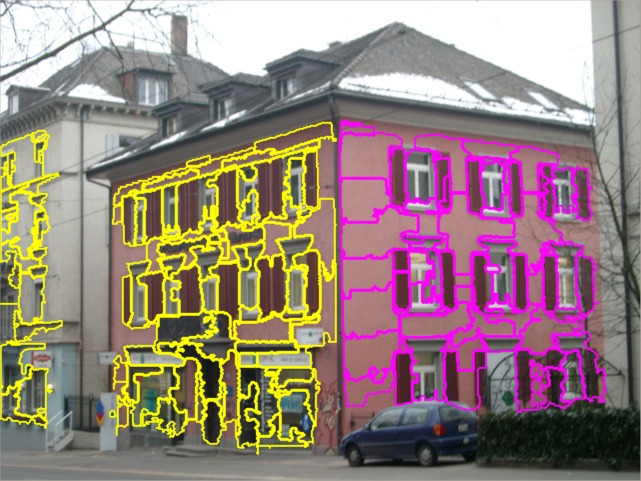} &
	\ic{0.09}{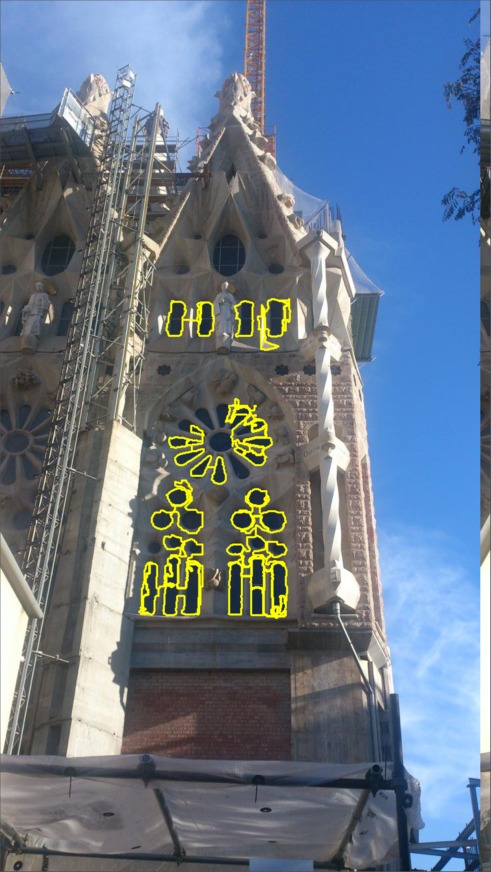} &
	\ic{0.215}{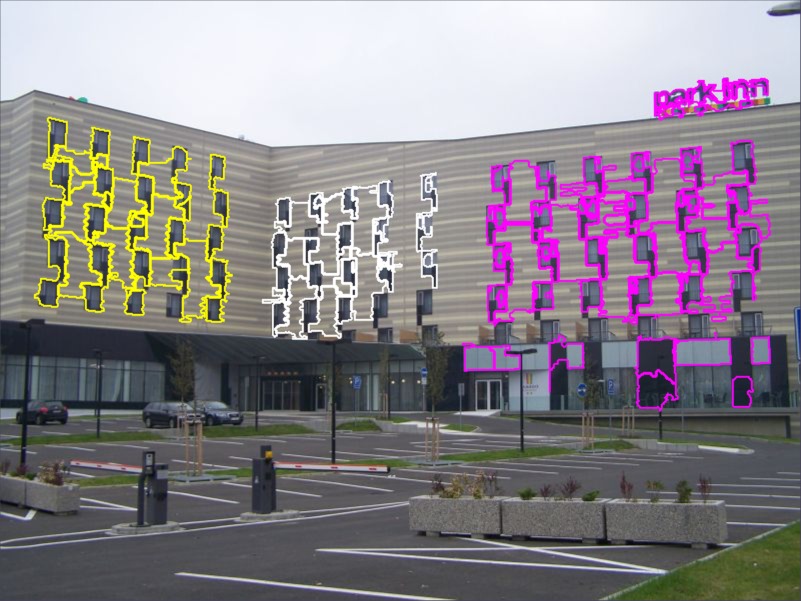} &
	\ic{0.243}{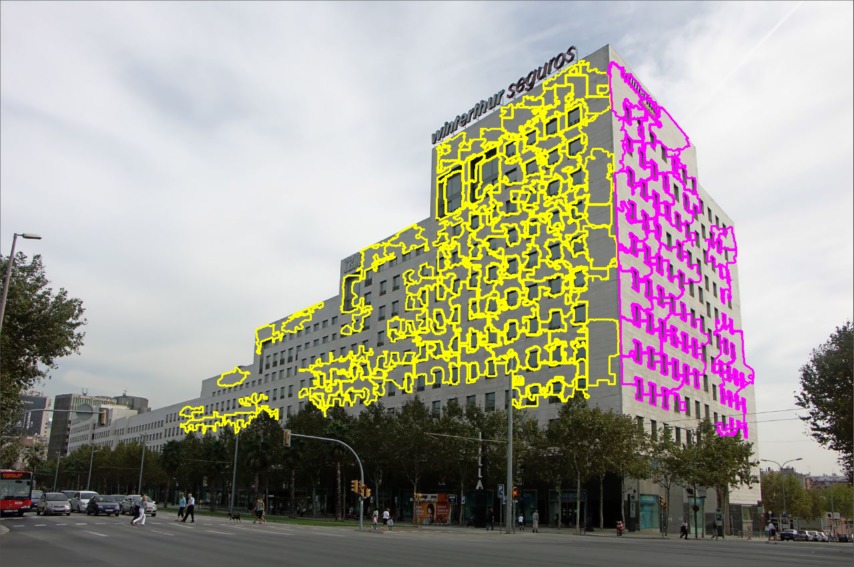} \\

    (h) &
	\ic{0.12}{img/img1_spixel_npoutlier.jpg} & 
	\ic{0.215}{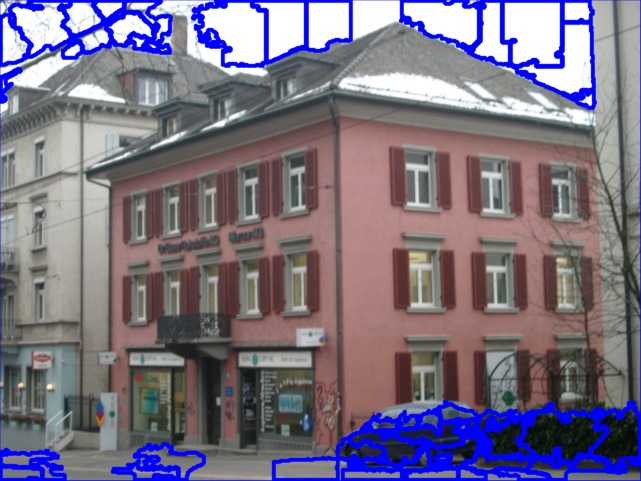} & 
	\ic{0.09}{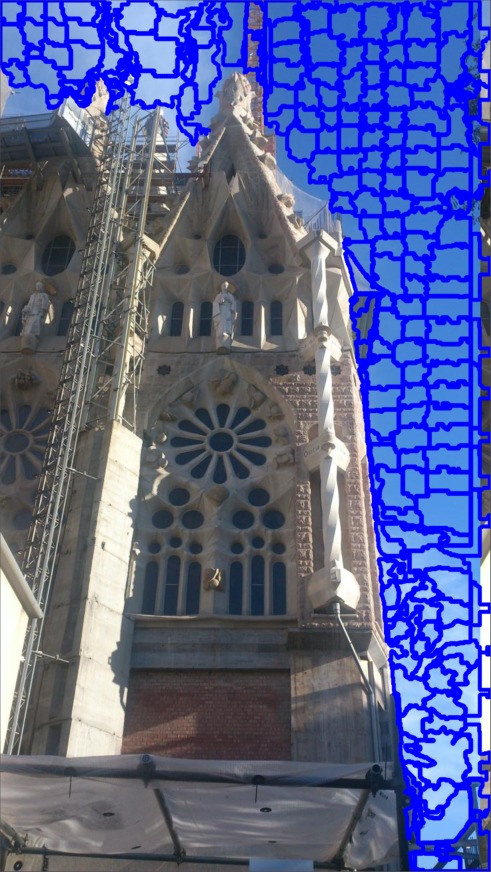} & 
	\ic{0.215}{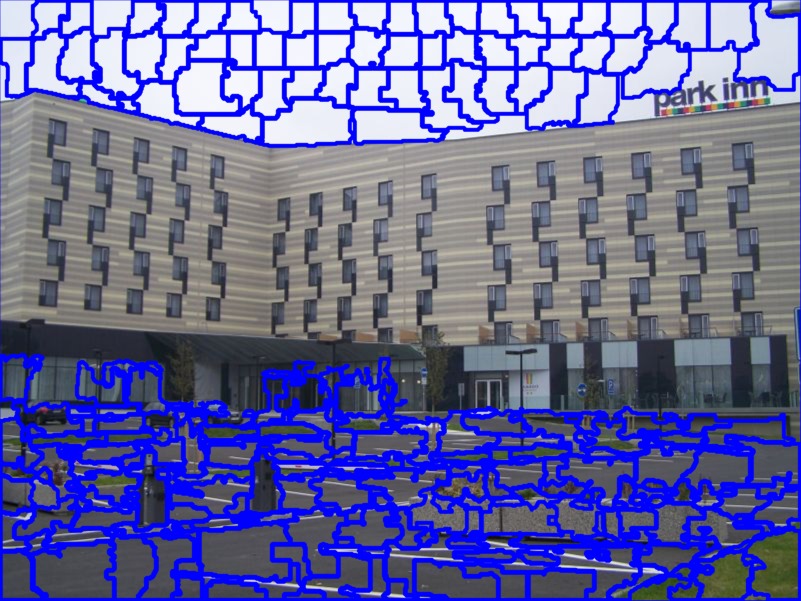} & 
	\ic{0.243}{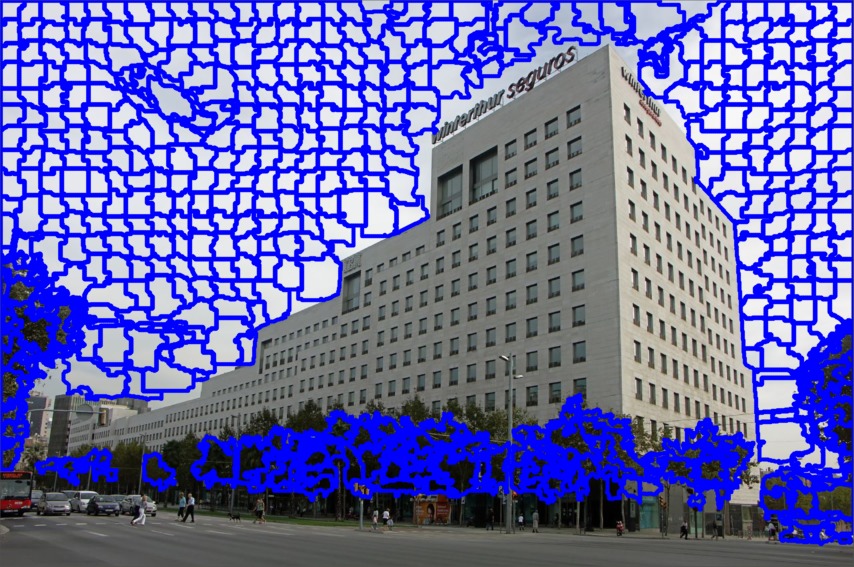}
\end{tabular}
\caption{Annotations: (a) constraints coplanar repeat grouping, (b)
  vanishing line assignment, (c) plane assignment, (d) mutually
  distinct repeated content, (e) coplanar repeats found by
  annotation-assisted inference, (f) features on the background
  surface, (g) vanishing line assignment for regions, (h) regions on
  the background surface.}
\end{figure}

%

\end{document}